\documentclass{article}

\usepackage{microtype}
\usepackage{graphicx}
\usepackage{booktabs} % for professional tables

\usepackage{hyperref}

\usepackage[preprint]{icml2026}

\usepackage{amsmath}
\usepackage{amssymb}
\usepackage{mathtools}
\usepackage{amsthm}
\usepackage{graphicx}
\usepackage{subcaption} 
\usepackage{multirow}

\usepackage[capitalize,noabbrev]{cleveref}

\theoremstyle{plain}

\theoremstyle{definition}

\theoremstyle{remark}

\usepackage[textsize=tiny]{todonotes}

\icmltitlerunning{Submission and Formatting Instructions for ICML 2026}

\begin{document}

\twocolumn[
  \icmltitle{Pairing Regularization for Mitigating Many-to-One Collapse in GANs}



  \icmlsetsymbol{equal}{*}

  \begin{icmlauthorlist}
    \icmlauthor{Kuan-Yu Lin}{nycu}
    \icmlauthor{Yu-Chih Huang}{nycu}
    \icmlauthor{Tie Liu}{tamu}
    % \icmlauthor{Firstname2 Lastname2}{equal,yyy,comp}
    % \icmlauthor{Firstname3 Lastname3}{comp}
  \end{icmlauthorlist}

  \icmlaffiliation{nycu}{Institute of Communications Engineering, National Yang Ming Chiao Tung University}
  \icmlaffiliation{tamu}{Department of Electrical \& Computer Engineering, Texas A\&M University}

  \icmlcorrespondingauthor{Kuan-Yu Lin}{casperlin.ee09@nycu.edu.tw}
  \icmlcorrespondingauthor{Yu-Chih Huang}{jerryhuang@nycu.edu.tw}
  \icmlcorrespondingauthor{Tie Liu}{tieliu@tamu.edu}

  \icmlkeywords{Machine Learning, ICML}

  \vskip 0.3in
]



\printAffiliationsAndNotice{}  % no special notice (required even if empty)

\begin{abstract}
Mode collapse remains a fundamental challenge in training generative adversarial networks (GANs). While existing works have primarily focused on inter-mode collapse, such as mode dropping, intra-mode collapse—where many latent variables map to the same or highly similar outputs—has received significantly less attention. In this work, we propose a pairing regularizer jointly optimized with the generator to mitigate the many-to-one collapse by enforcing local consistency between latent variables and generated samples.

We show that the effect of pairing regularization depends on the dominant failure mode of training. In collapse-prone regimes with limited exploration, pairing encourages structured local exploration, leading to improved coverage and higher recall. In contrast, under stabilized training with sufficient exploration, pairing refines the generator’s induced data density by discouraging redundant mappings, thereby improving precision without sacrificing recall. Extensive experiments on both toy distributions and real-image benchmarks demonstrate that the proposed regularizer effectively complements existing stabilization techniques by directly addressing intra-mode collapse.
\end{abstract}

% \begin{figure}[t]
%     \centering
%     \begin{subfigure}{0.48\linewidth}
%         \centering
%         \includegraphics[width=\linewidth]{pr_25M_scatter.pdf}
%         \caption{Precision--recall comparison at 25M images.}
%         \label{fig:pr_final}
%     \end{subfigure}
%     \hfill
%     \begin{subfigure}{0.48\linewidth}
%         \centering
%         \includegraphics[width=\linewidth]{pr_trajectory_seed0_25M.png}
%         \caption{PR trajectory over training (3M--25M).}
%         \label{fig:pr_trajectory}
%     \end{subfigure}

%     \caption{
%     \textbf{Precision--recall performance on CIFAR-10 with ADA.}
%     (\textbf{Left}) Final precision--recall comparison at 25M training images.
%     (\textbf{Right}) Precision--recall trajectory over training.
%     Pairing regularization consistently improves recall while maintaining comparable or higher precision,
%     indicating enhanced mode coverage and altered training dynamics.
%     }
%     \label{fig:pr_main}
% \end{figure}

\section{Introduction}

\begin{figure*}[t]
    \centering

    \begin{minipage}[t]{0.23\textwidth}
        \centering
        \includegraphics[width=0.9\linewidth]{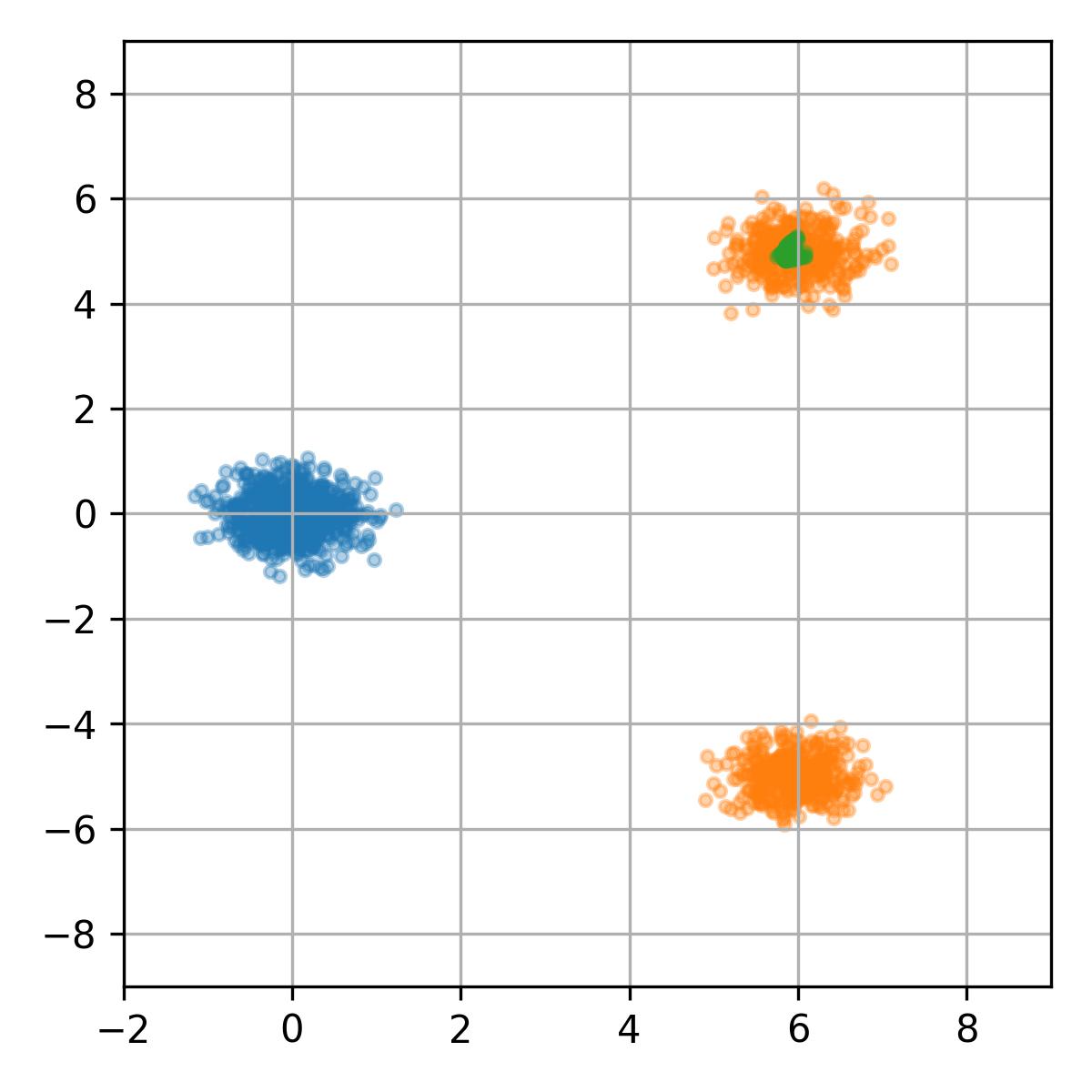}
        \\ (a) GAN
    \end{minipage}
    \hspace{0.01\textwidth}
    \begin{minipage}[t]{0.23\textwidth}
        \centering
        \includegraphics[width=0.9\linewidth]{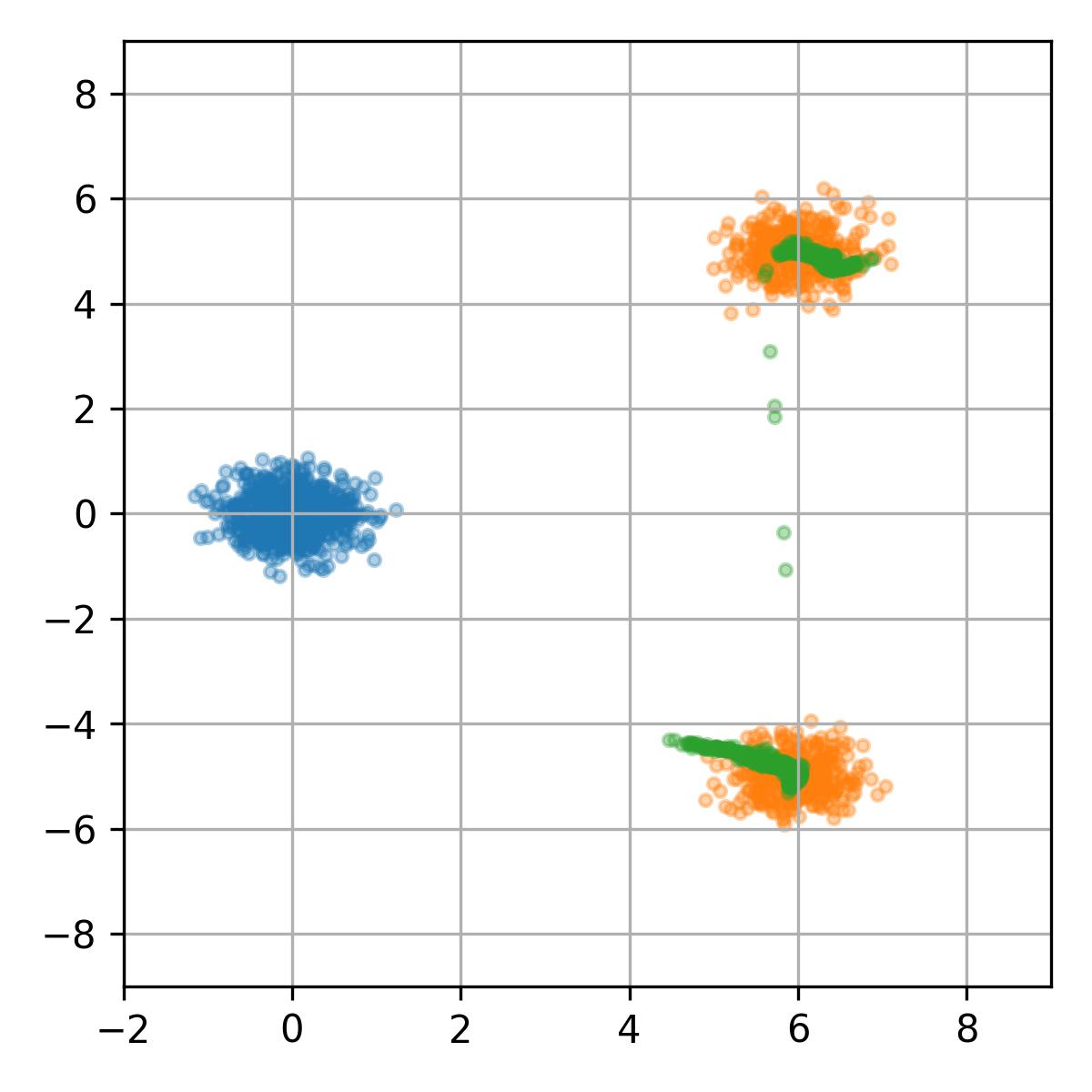}
        \\ (b) GAN + R1
    \end{minipage}
    \hspace{0.01\textwidth}
    \begin{minipage}[t]{0.23\textwidth}
        \centering
        \includegraphics[width=0.9\linewidth]{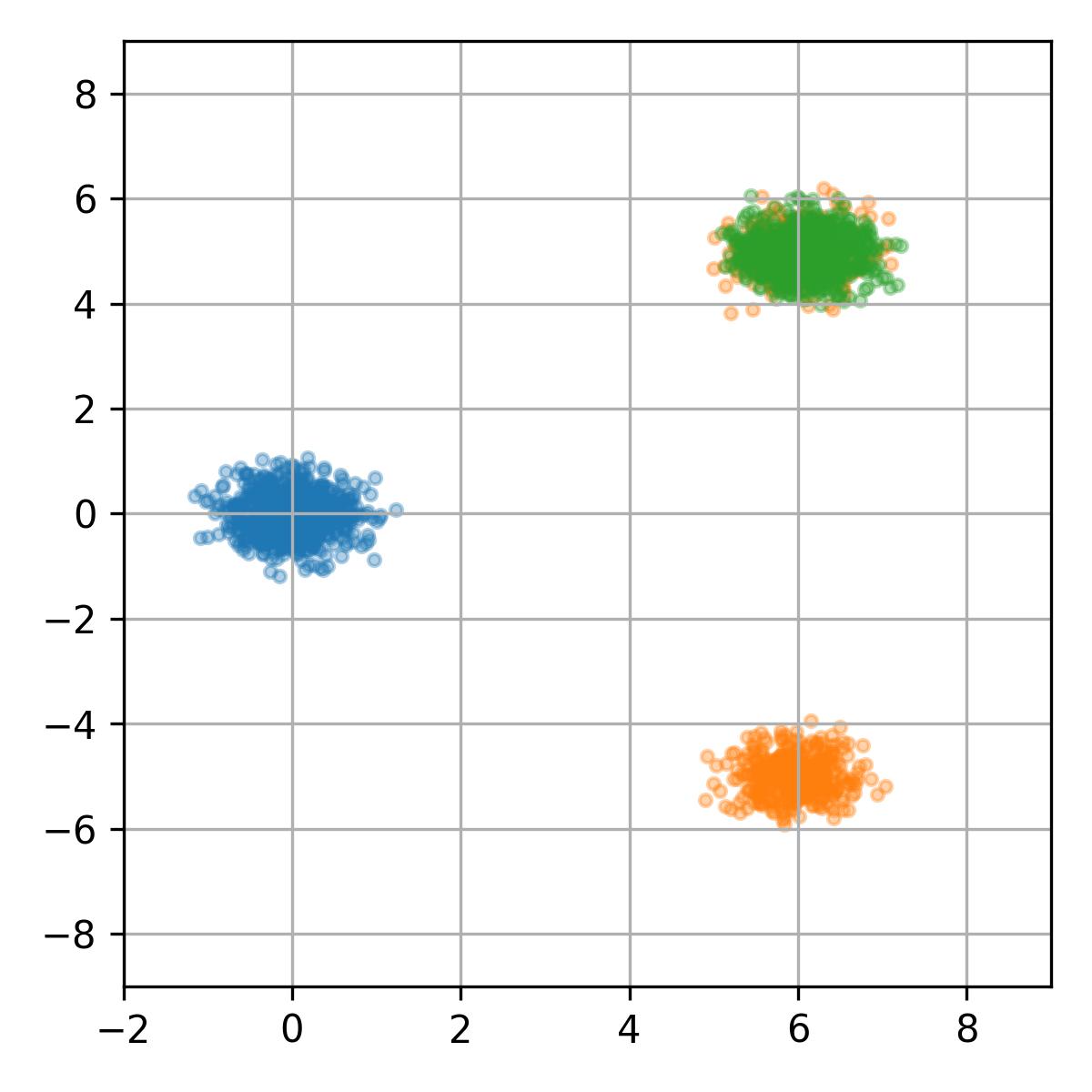}
        \\ (c) GAN + Pairing
    \end{minipage}
    \hspace{0.01\textwidth}
    \begin{minipage}[t]{0.23\textwidth}
        \centering
        \includegraphics[width=0.9\linewidth]{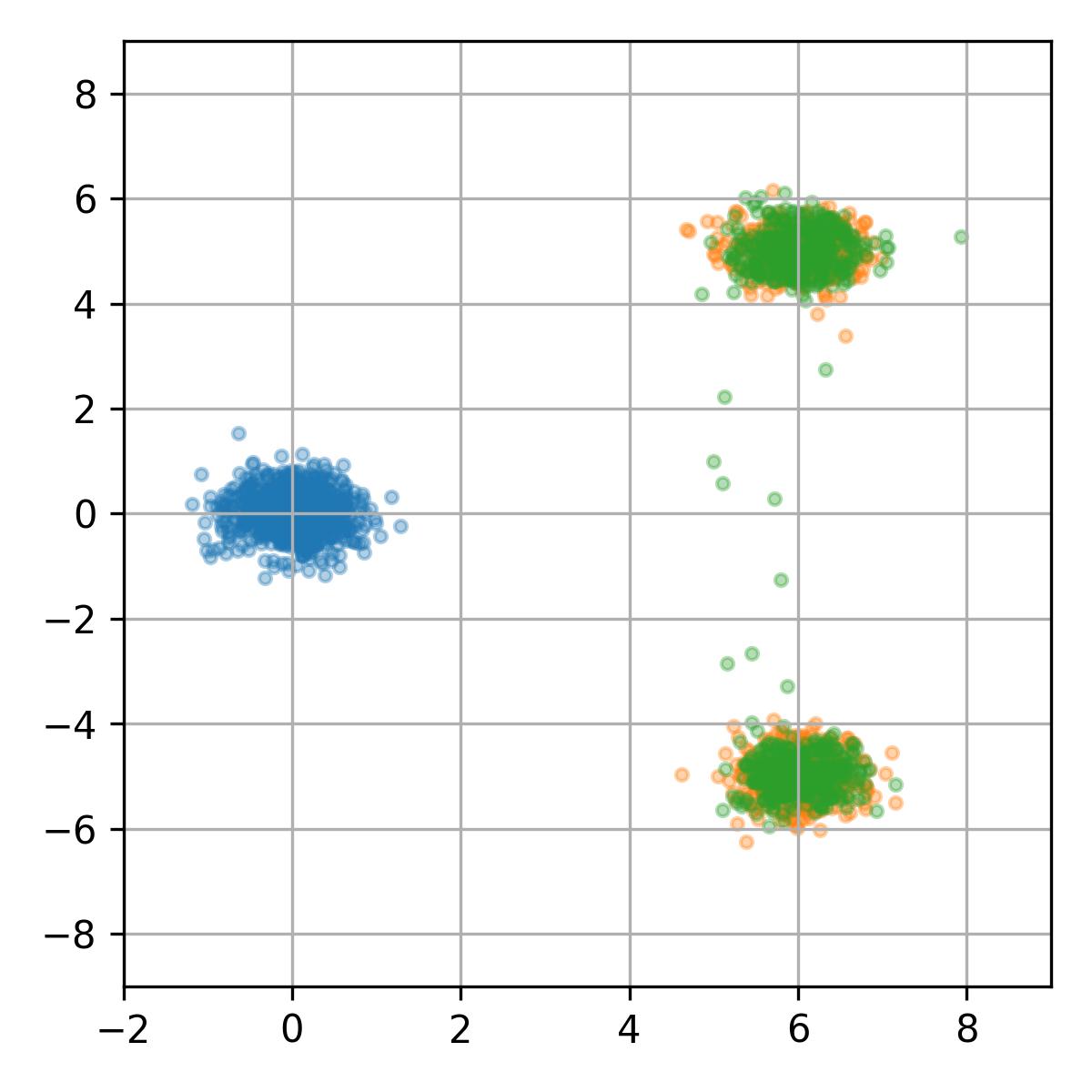}
        \\ (d) GAN + Pairing + R1
    \end{minipage}

    \caption{
    Two-dimensional Gaussian mixture illustrating inter- and intra-mode collapse.
    (a) Standard GAN exhibits inter-mode collapse, failing to cover all mixture components.
    (b) Adding gradient penalty (R1) avoids mode dropping but results in highly concentrated samples within each mode, indicating persistent many-to-one (intra-mode) collapse.
    (c) Pairing regularization increases intra-mode diversity and mitigates many-to-one collapse, though some modes remain uncovered.
    (d) Combining pairing regularization with gradient penalty addresses both failure modes, achieving improved mode coverage and well-distributed samples within each mode.
    }
    \label{fig:2component_gaussian}
\end{figure*}

\begin{figure*}[t]
    \centering

    \begin{minipage}[t]{0.23\textwidth}
        \centering
        \includegraphics[width=0.9\linewidth]{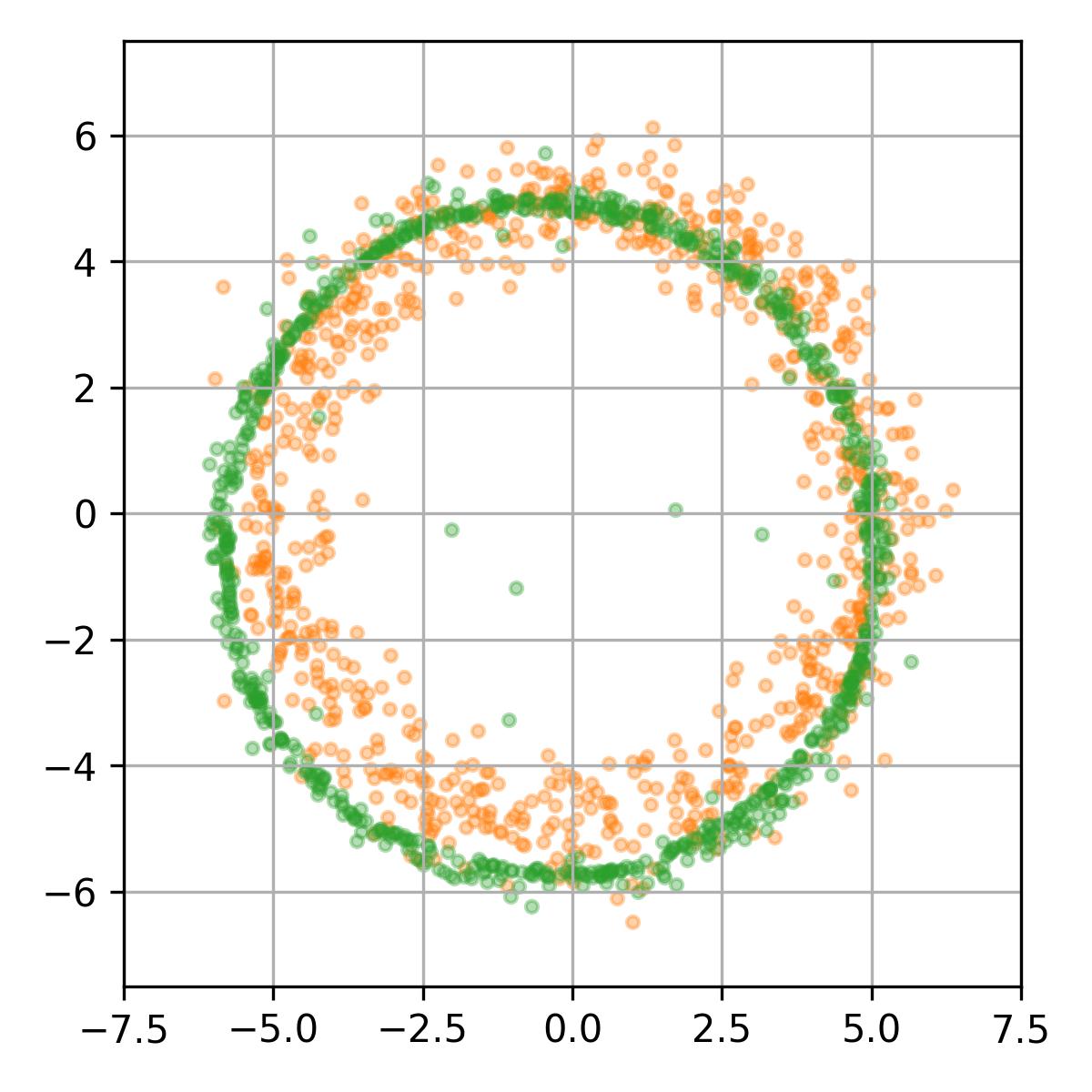}
        \\ (a) GAN
    \end{minipage}
    \hspace{0.01\textwidth}
    \begin{minipage}[t]{0.23\textwidth}
        \centering
        \includegraphics[width=0.9\linewidth]{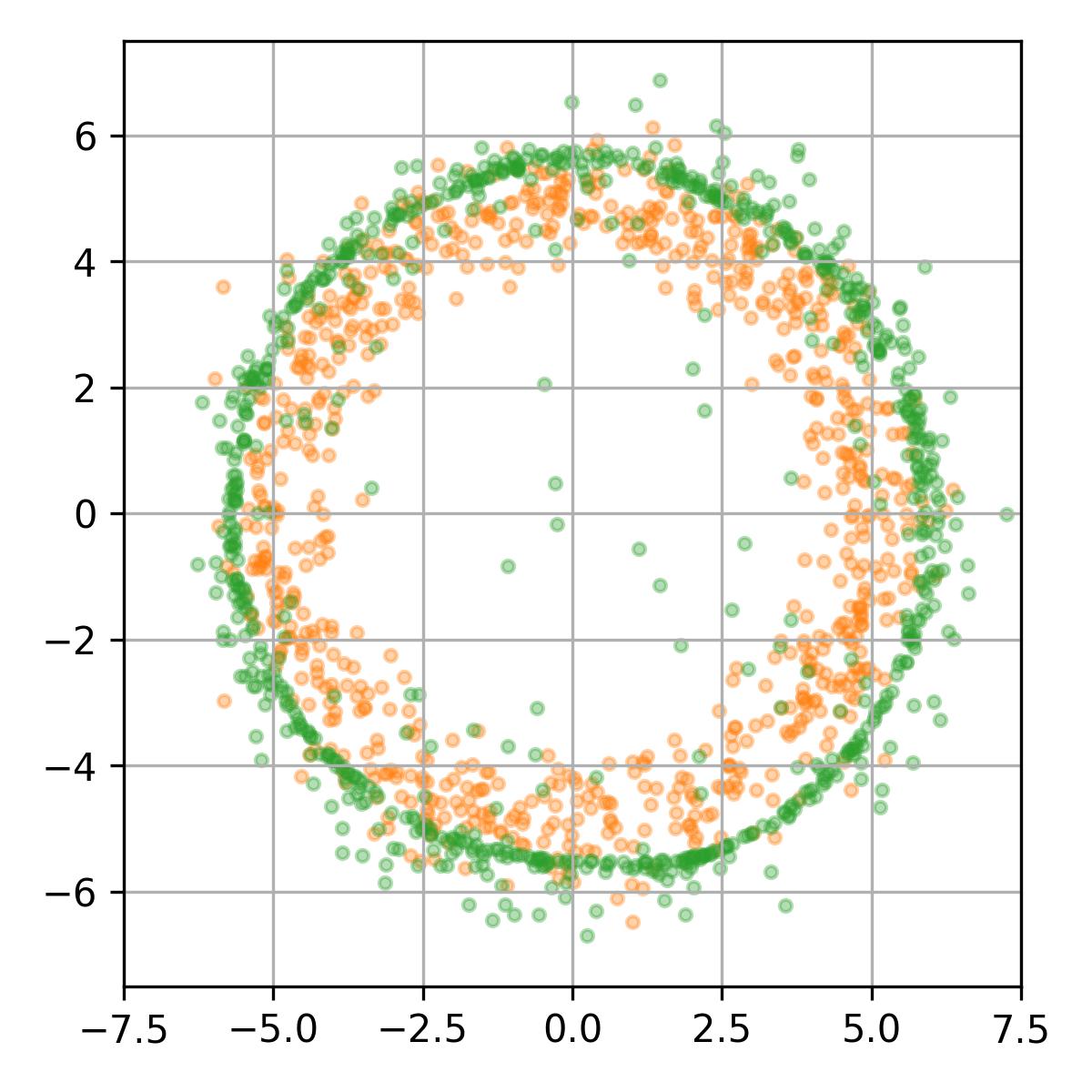}
        \\ (b) GAN + R1
    \end{minipage}
    \hspace{0.01\textwidth}
    \begin{minipage}[t]{0.23\textwidth}
        \centering
        \includegraphics[width=0.9\linewidth]{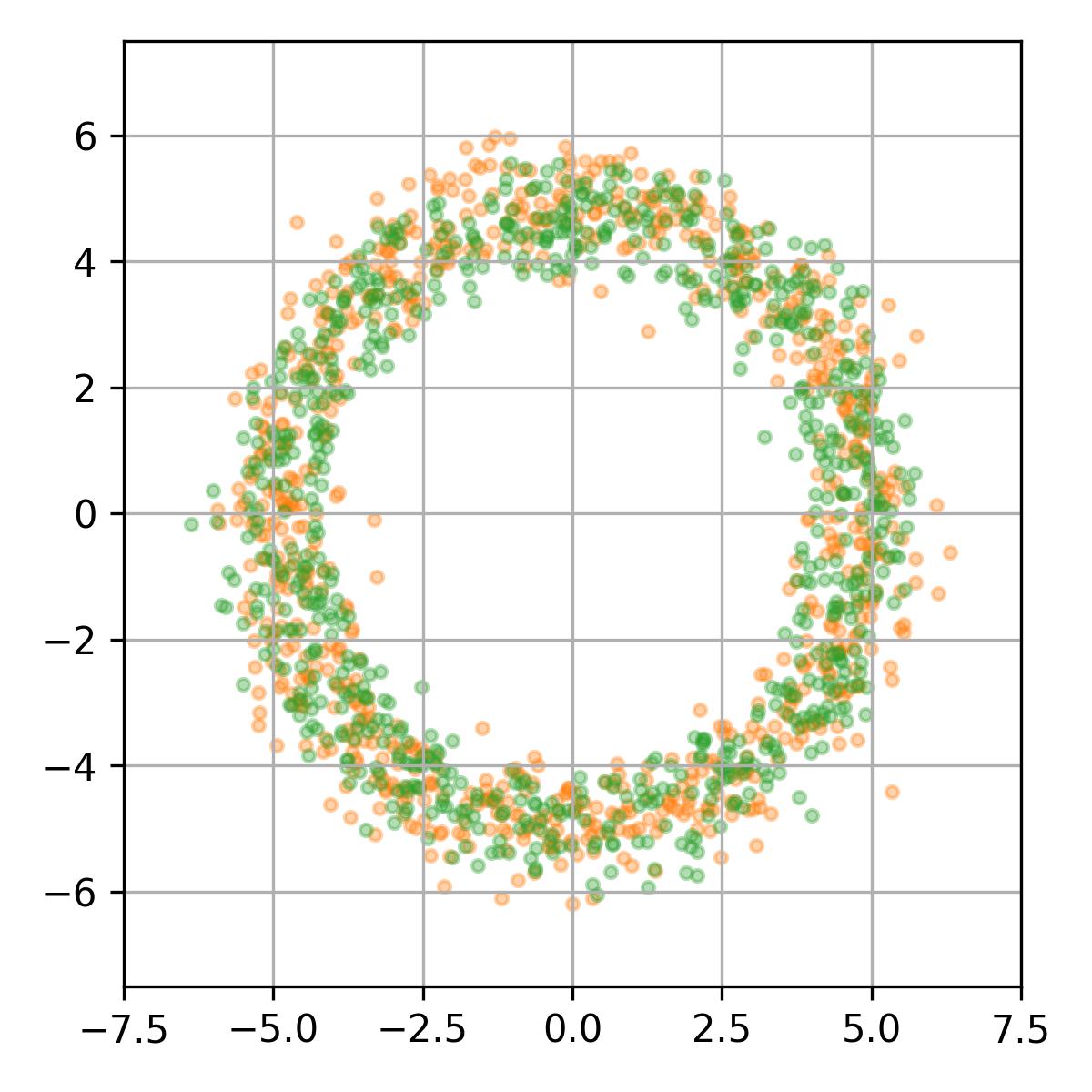}
        \\ (c) GAN + Pairing
    \end{minipage}
    \hspace{0.01\textwidth}
    \begin{minipage}[t]{0.23\textwidth}
        \centering
        \includegraphics[width=0.9\linewidth]{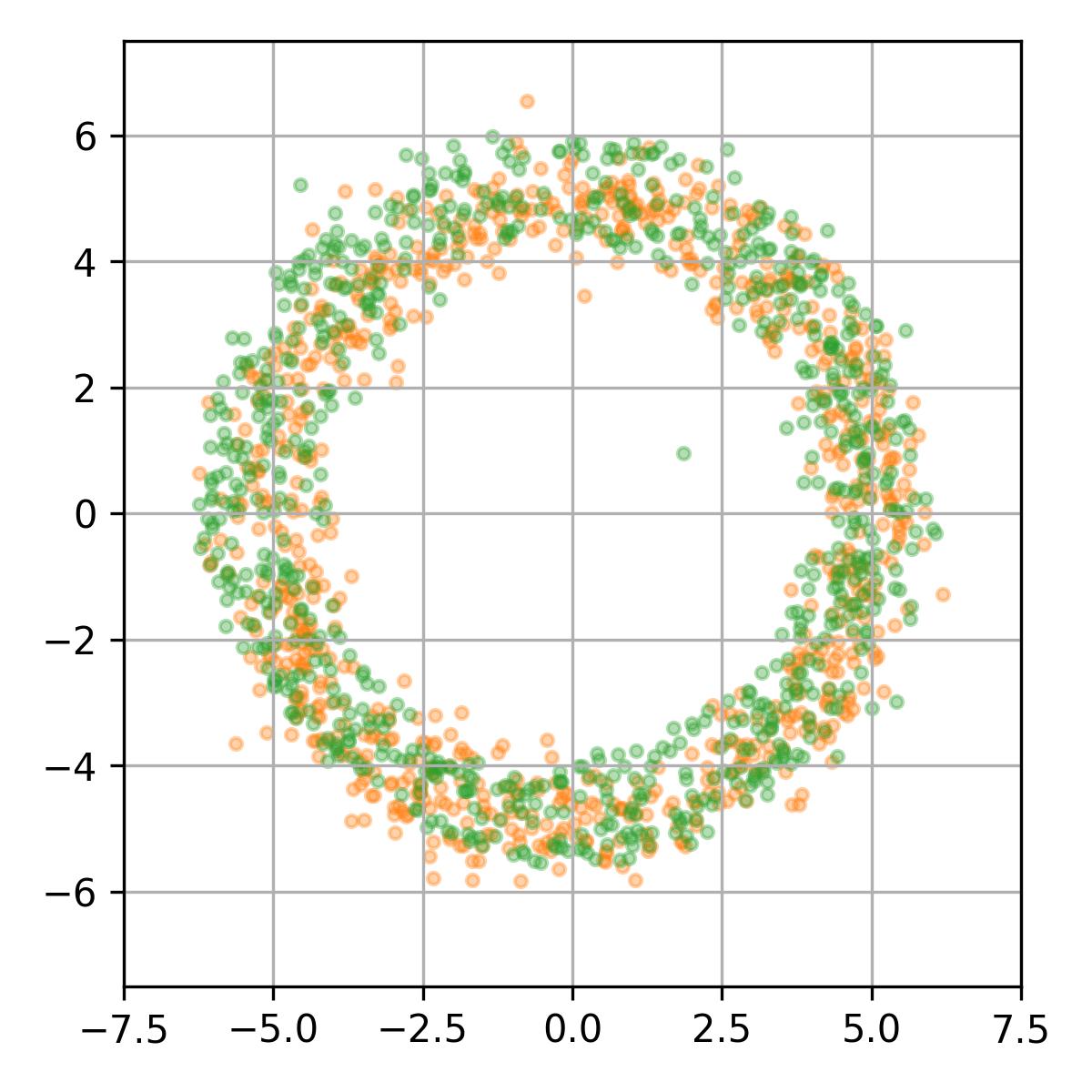}
        \\ (d) GAN + Pairing + R1
    \end{minipage}

    \caption{
    Two-dimensional ring distribution illustrating many-to-one collapse in high-recall regime. All methods achieve similar coverage of the data manifold, yet differ in how probability mass is distributed along the ring. (a) Standard GAN produces uneven and concentrated samples, indicating many-to-one collapse. (b) Gradient penalty improves training stability but does not fully resolve intra-mode collapse.
    (c) Pairing regularization redistributes samples more uniformly along the manifold.
    (d) The combination of pairing regularization and gradient penalty yields stable training and improved coverage of the manifold.
    }
    \label{fig:ring}
\end{figure*}

Modern generative adversarial networks (GANs)~\cite{karras2020stylegan2ada, zhao2020differentiableaugmentation, kang2023scaling,takida2023san, huang2024r3gan} have achieved remarkable progress through improved training stability and regularization techniques. Methods such as gradient penalties~\cite{mescheder2018gp,mescheder2018which}, spectral normalization~\cite{miyato2018spectralnorm}, and data augmentation~\cite{karras2020stylegan2ada,zhao2020differentiableaugmentation} have substantially mitigated mode dropping, enabling generators to cover a larger portion of the data manifold. Consequently, recall-based metrics~\cite{sajjadi2018assessing,kynkaanniemi2019improved} are often used as primary indicators of training success, and high recall is frequently interpreted as evidence that the generator has learned the target distribution.

However, improved training stability alone does not guarantee that the generator learns a faithful mapping from latent space to data space~\cite{arjovsky2017wgan,donahue2016adversarial,kynkaanniemi2019improved,karras2020analyzing}. Even when mode dropping is mitigated, the generator may still fail to preserve meaningful variation within each mode. In particular, large and distinct regions of the latent space can be mapped to highly overlapping regions in data space, resulting in collapsed structure within modes despite apparently stable training. This form of collapse is qualitatively different from mode dropping. Rather than missing modes altogether, the generator collapses variation within modes, producing samples that lack internal diversity~\cite{bau2019seeing}. As illustrated in Figure~\ref{fig:2component_gaussian}(a) and Figure~\ref{fig:2component_gaussian}(b), gradient penalties substantially improve mode coverage, yet samples within each mode remain highly concentrated, indicating that training stabilization alone is insufficient to prevent this failure.

We refer to this failure mode as many-to-one collapse. Unlike mode dropping, many-to-one collapse arises when multiple, distinct latent regions are mapped to the same or highly overlapping regions in data space, leading to a loss of variation within modes. Crucially, this form of collapse can persist even under stable training and extensive regularization, where conventional indicators suggest convergence. In such regimes, recall-based metrics become insufficient for diagnosing generator quality. While recall measures whether the generator reaches different regions of the data manifold, it remains insensitive to how probability mass is distributed within those regions. This limitation motivates the use of complementary diagnostics, such as coverage, which quantify how generated samples populate the data manifold beyond mere support coverage. We illustrate this phenomenon in Figure \ref{fig:ring} and Table~\ref{tab:ring_pr_coverage}: although all methods achieve high recall on a circular distribution, the resulting sample distributions differ substantially in their coverage of the manifold, revealing pronounced concentration of probability mass under standard training and substantially more uniform coverage when many-to-one collapse is mitigated.

These observations suggest that many-to-one collapse cannot be addressed solely through improved exploration or training stabilization. Rather, it arises from a missing structural constraint on the generator mapping—one that governs how variations in latent space are translated into variations in data space. When such a constraint is absent, large latent regions may collapse onto similar outputs without being penalized by standard adversarial objectives.

\begin{table}[t]
\centering
\caption{Quantitative results on the two-dimensional Gaussian mixture
corresponding to Figure~1.
Combining pairing regularization with gradient
penalty achieves high recall and high coverage~\cite{naeem2020prdc}.}
\label{tab:2component_pr}
\small
\setlength{\tabcolsep}{6pt} 
\begin{tabular}{lccc}
\toprule
Method & Precision & Recall  & Coverage \\
\midrule
GAN              & 0.992 & 0.184 & 0.083 \\
GAN + R1      & 0.967 & 0.267 & 0.283 \\
GAN + Pairing    & 0.974 & 0.494 & 0.483\\
GAN + Pairing + R1 & 0.969 & \textbf{0.957} & \textbf{0.861} \\
\bottomrule
\end{tabular}
\end{table}
\begin{table}[t]
\centering
\caption{Quantitative results on the two-dimensional ring distribution corresponding to Figure~\ref{fig:ring}.
While gradient penalty (R1) improves recall, it fails to recover coverage in high-recall regimes. Pairing regularization substantially improves coverage.}
\label{tab:ring_pr_coverage}
\small
\setlength{\tabcolsep}{6pt}   % default ~6pt, can go to 5pt if needed
\begin{tabular}{lccc}
\toprule
Method & Precision $\uparrow$ & Recall $\uparrow$ & Coverage $\uparrow$ \\
\midrule
GAN & 0.981 & 0.754 & 0.336 \\
GAN + R1 & 0.907 & 0.948 & 0.372 \\
GAN + Pairing & 0.970 & \textbf{0.962} & \textbf{0.851} \\
GAN + Pairing + R1 & 0.975 & 0.944 & 0.841 \\
\bottomrule
\end{tabular}
\end{table}
To address this structural deficiency, we introduce pairing regularization as a complementary constraint on the generator mapping. Pairing regularization explicitly encourages local correspondence between neighborhoods in latent space and their generated outputs, thereby directly targeting many-to-one collapse. Moreover, this regularization is orthogonal to existing stabilization techniques and can be seamlessly combined with gradient penalties and other regularization strategies. The effect of pairing regularization is consistently reflected in Figures~\ref{fig:2component_gaussian} and Figure~\ref{fig:ring}. As shown in Figure~\ref{fig:2component_gaussian}, pairing alleviates the concentration of samples within modes that persists under gradient penalties alone. Figure~\ref{fig:ring} further shows that pairing leads to substantially improved coverage of the data manifold, even when all methods achieve similarly high recall. These results show that pairing regularization addresses a failure mode that training stabilization alone cannot resolve.

We validate the proposed pairing regularization through a series of controlled experiments and real-data evaluations. In addition to the illustrative two-dimensional examples, we study GAN training on CIFAR-10 under both standard and stabilized training settings, demonstrating that the failure modes highlighted in Figures 1 and 2 manifest in practical scenarios as well. Across these settings, pairing regularization consistently improves the alignment between latent variation and generated samples, complementing existing stabilization techniques.

In summary, this work makes three contributions. First, we identify many-to-one collapse as a failure mode distinct from mode dropping, which persists even under stable training and high recall. Second, we introduce pairing regularization as a complementary constraint on the generator mapping that directly targets this failure. Third, we provide empirical evidence across diagnostic toy examples and CIFAR-10 evaluations, highlighting the limitations of recall-based evaluation alone and the effectiveness of pairing in mitigating many-to-one collapse.

\section{Backgroud}\label{sec:background}
\subsection{Inter-Mode and Intra-Mode Collapse}
Mode collapse is a long-standing challenge in generative adversarial networks~\cite{goodfellow2020generative,arjovsky2017wgan,shahbazi2022collapse}. A commonly studied form of collapse is mode dropping, where the generator fails to cover all modes of the target data distribution, producing samples from only a subset of the data manifold. This phenomenon has been extensively investigated, and numerous techniques have been proposed to mitigate it by improving exploration and training stability.

However, even when mode coverage is achieved, generators may still exhibit a different form of collapse. In this case, although all modes of the data distribution are reached, the generator fails to preserve meaningful variation within each mode. Concretely, multiple and distinct regions of the latent space may be mapped to the same or highly overlapping regions in data space, resulting in limited intra-mode diversity despite apparent coverage of the data manifold.

This distinction between inter-mode collapse (mode dropping) and intra-mode collapse is not merely semantic. The two failure modes manifest differently in practice and respond to different forms of regularization. While inter-mode collapse concerns whether the support of the generated distribution covers that of the data, intra-mode collapse concerns how probability mass is distributed within the covered regions. As a result, a generator may appear successful under conventional coverage-based criteria while still exhibiting severe structural deficiencies in its mapping from latent space to data space.
\subsection{Limitations of Stabilization-Based Training}
Recent advances in GAN training have substantially improved stability and mitigated mode dropping~\cite{mescheder2018which,miyato2018spectralnorm,karras2020stylegan2ada,zhao2020differentiableaugmentation}. Despite their effectiveness, stabilization-based techniques primarily act on the discriminator or on the overall training dynamics. By design, they aim to regularize optimization behavior—encouraging smooth gradients, preventing exploding updates, and stabilizing adversarial training. However, they do not explicitly constrain the structure of the generator mapping itself, nor do they regulate how variations in latent space should be translated into variations in generated samples. 

Even under stable training and extensive regularization, large regions of the latent space may collapse onto similar outputs without being penalized by standard adversarial objectives. This limitation motivates the need for complementary constraints that operate directly on the generator mapping, addressing intra-mode collapse at its structural source rather than further stabilizing training dynamics alone.

\subsection{Limitations of Diversity-Based Methods}
Beyond stabilization-based training, a complementary line of work~\cite {chen2016infogan,mao2019mode,yang2019diversity,pan2022unigan,allahyani2023divgan,gong2024testing} introduces diversity-oriented objectives to mitigate collapse. Representative approaches encourage output dispersion by maximizing distances between generated samples~\cite{mao2019mode}, or by promoting mutual information between structured latent codes and the generated data to encourage disentangled factors of variation ~\cite{chen2016infogan}. These methods have been shown to improve distribution-level coverage and alleviate classical mode dropping in practice, but do not explicitly constrain the correspondence between latent variables and generated samples.

However, such objectives typically operate at an aggregate or distributional level, without enforcing explicit correspondence between individual latent variables and generated samples. They may increase overall sample diversity while still allowing many-to-one mappings within local regions of the latent space. In particular, distinct latent codes can remain indistinguishable under the generator mapping as long as sufficient global variation is preserved.

This limitation highlights a gap between promoting diversity in marginal sample distributions and preserving structural fidelity in the latent-to-data mapping. Addressing intra-mode collapse therefore requires constraints that act directly on the generator mapping at a finer granularity, beyond improving training stability or encouraging global diversity alone.

\section{Pairing Regularization}
We consider a standard GAN consisting of a generator trained under an adversarial objective. 
Our focus in this work is exclusively on the generator mapping $G$, and we do not modify 
the discriminator architecture or the adversarial loss. 
Pairing regularization is introduced as an additional constraint on the generator 
and can be applied on top of existing GAN formulations and stabilization techniques.

\subsection{Intuition}
As discussed in Section~\ref{sec:background}, many-to-one collapse reflects a structural deficiency in the generator mapping rather than insufficient exploration or unstable training. In particular, when large and distinct regions of the latent space are mapped to highly overlapping regions in data space, the generator fails to preserve meaningful variation within modes—even when mode coverage is achieved.

Conceptually, this failure can be understood as a breakdown in local correspondence between the latent and data spaces. Ideally, nearby points in latent space should map to nearby yet distinct samples in data space, so that local variations in the latent input are reflected in the generated outputs. Many-to-one collapse occurs when this correspondence is violated: neighborhoods in latent space map to nearly identical outputs, leading to loss of variation without being penalized by the adversarial objective. 

Pairing regularization is motivated by the need to explicitly encourage this local correspondence. Rather than modifying the discriminator or further stabilizing the training dynamics, pairing acts directly on the generator by promoting consistency between latent-space and generated-sample variations.

\subsection{Loss Formulation}\label{sec:pairing_loss}
To enforce the pairing of latent variables with generated samples, we define this pairing as a latent identification problem within each minibatch.
Given latent variables $\{z_i\}_{i=1}^B$ sampled from the prior and the corresponding generated samples
$G_\theta(z_i)$,
the objective is to correctly associate each generated sample $G_\theta(z_i)$ with its originating latent variable $z_i$
among the set of candidates $\{z_j\}_{j=1}^B$.
When the generator suffers from many-to-one collapse, this identification task becomes ambiguous,
as distinct latent variables produce nearly indistinguishable outputs.

A simple instantiation of the latent identification objective is to construct mismatched pairs
by randomly permuting latent variables within a minibatch.
Let $\pi$ denote a random permutation of $\{1,\dots,B\}$ such that $\pi(i)\neq i$,
and define mismatched pairs $(G_\theta(z_i), z_{\pi(i)})$.
A binary classifier can then be trained to distinguish matched pairs $(G_\theta(z_i), z_i)$
from mismatched pairs $(G_\theta(z_i), z_{\pi(i)})$.
While intuitive, this formulation uses only a single negative example per sample
and does not fully exploit the rich set of mismatches available within the minibatch.

To more effectively enforce pairability, we adopt a contrastive identification formulation
that contrasts each generated sample against mismatched latent variables in the minibatch. For each generated sample $G_\theta(z_i)$, we classify which latent variable in $\{z_j\}_{j=1}^B$ produced it.
Specifically, we define logits $\ell_{ij} = s_{ij}/\tau$ with temperature $\tau>0$,
and minimize the following pairing loss,
\begin{equation}
\mathcal{L}_{\text{pair}}
=
\frac{1}{B}\sum_{i=1}^B
\left[
- \log
\frac{\exp(\ell_{ii})}{\sum_{j=1}^B \exp(\ell_{ij})}
\right].
\label{eq:pair_loss}
\end{equation}

Although the pairing loss takes a contrastive form,
its role here is not representation learning.
Rather, it serves as a structural regularizer on the generator mapping.
Gradients from $\mathcal{L}_{\text{pair}}$ are backpropagated into the generator,
encouraging distinct latent variables to produce distinguishable samples.
This directly penalizes many-to-one mappings and enforces pairability at a local scale,
complementing adversarial training objectives that operate at the distribution level.

\begin{algorithm}[t]
\caption{Training Pairing-Regularized GAN}
\label{alg:pairing_gan}
\begin{algorithmic}[1]
\REQUIRE
Generator $G_\theta$, discriminator $D_\psi$, pairing network $P_\phi$;
pairing weight $\lambda_{\text{pair}}$;
latent prior $p(z)$;
data distribution $p_{\text{data}}$

\WHILE{not converged}
    \STATE Sample minibatch $\{x_i\}_{i=1}^B \sim p_{\text{data}}$
    \STATE Sample minibatch $\{z_i\}_{i=1}^B \sim p(z)$

    \STATE \textbf{Discriminator update:}
    \STATE \hspace{0.5cm} Update $D_\psi$ using the standard GAN loss
    \STATE \hspace{0.5cm} Optionally apply discriminator regularization %(e.g., R1/R2)

    \STATE \textbf{Generator forward pass:}
    \STATE \hspace{0.5cm} Generate samples $x_i^{\text{gen}} = G_\theta(z_i)$

    \STATE \textbf{Pairing construction:}
    \STATE \hspace{0.5cm} Form matched pairs $(x_i^{\text{gen}}, z_i)$
    \STATE \hspace{0.5cm} Construct mismatched pairs $(x_i^{\text{gen}}, z_{\pi(i)})$
    \STATE \hspace{0.5cm} where $\pi$ is a random permutation with $\pi(i) \neq i$
    \STATE \hspace{0.5cm} Compute pairing loss $\mathcal{L}_{\text{pair}}$ %(e.g., binary or InfoNCE contrastive objective)
    \STATE \textbf{Generator and pairing update:}
    \STATE \hspace{0.5cm} Compute GAN generator loss $\mathcal{L}_{\text{GAN}}$
    \STATE \hspace{0.5cm} Update $\theta, \phi$ by minimizing $\mathcal{L}_{\text{GAN}} + \lambda_{\text{pair}} \mathcal{L}_{\text{pair}}$
\ENDWHILE
\end{algorithmic}
\end{algorithm}
\subsection{Training Objective and Optimization}\label{sec:training}

We consider a standard adversarial training objective augmented with the proposed pairing regularization.
The overall optimization problem is formulated as,
\begin{equation}
\min_{\theta,\phi}\;\max_{\psi}\;
\mathcal{L}_{\text{GAN}}(G_\theta, D_\psi)
+
\lambda_{\text{pair}}\,\mathcal{L}_{\text{pair}}(G_\theta, P_\phi),
\label{eq:overall_obj}
\end{equation}
where $G_\theta$ and $D_\psi$ denote the generator and discriminator, respectively,
$P_\phi$ denotes the pairing network,
and $\lambda_{\text{pair}}$ controls the strength of the pairing regularization. We optimize the above objective by alternately updating the discriminator
and jointly updating the generator and pairing network.
The complete training procedure is summarized in Algorithm~\ref{alg:pairing_gan}.

\paragraph{Generator-side update.}
The generator and the pairing network are jointly optimized.
Specifically, the generator parameters $\theta$ and pairing network parameters $\phi$
are updated to minimize the following objective,
\begin{equation}
\mathcal{L}_{\text{gen}}
=
\mathcal{L}_{\text{GAN}}^{(G)}(G_\theta, D_\psi)
+
\lambda_{\text{pair}}\,\mathcal{L}_{\text{pair}}(G_\theta, P_\phi),
\label{eq:gen_obj}
\end{equation}
where $\mathcal{L}_{\text{GAN}}$ denotes the standard adversarial loss
and $\mathcal{L}_{\text{pair}}$ is the pairing loss defined in Section~3.3.
Gradients from the pairing loss are backpropagated only into the generator and pairing network.

\paragraph{Discriminator-side update.}
The discriminator is trained independently by minimizing the following objective,
\begin{equation}
\mathcal{L}_{\text{disc}}
=
\mathcal{L}_{\text{GAN}}^{(D)}(G_\theta, D_\psi),
\label{eq:disc_obj}
\end{equation}
where $\mathcal{L}_{\text{GAN}}^{(D)}$ denotes the discriminator-side adversarial loss. For training stabilization, standard regularization techniques such as a zero-centered gradient penalty
can be optionally added to the discriminator objective when appropriate.

\section{Experiments}

We evaluate the proposed pairing-regularized GAN on a range of synthetic and
real-world benchmarks to assess its ability to mitigate mode collapse and improve
latent--sample correspondence. We first use
toy 2D Gaussian mixtures to visualize many-to-one collapse and its mitigation, then quantify mode coverage in a high-recall regime, and finally evaluate scalability and
sample quality on CIFAR-10~\cite{krizhevsky2009cifar} using precision--recall metrics.
\begin{figure}[t]
    
    \centering
    \begin{minipage}[t]{0.46\linewidth}
        \centering
        \includegraphics[width=\linewidth]{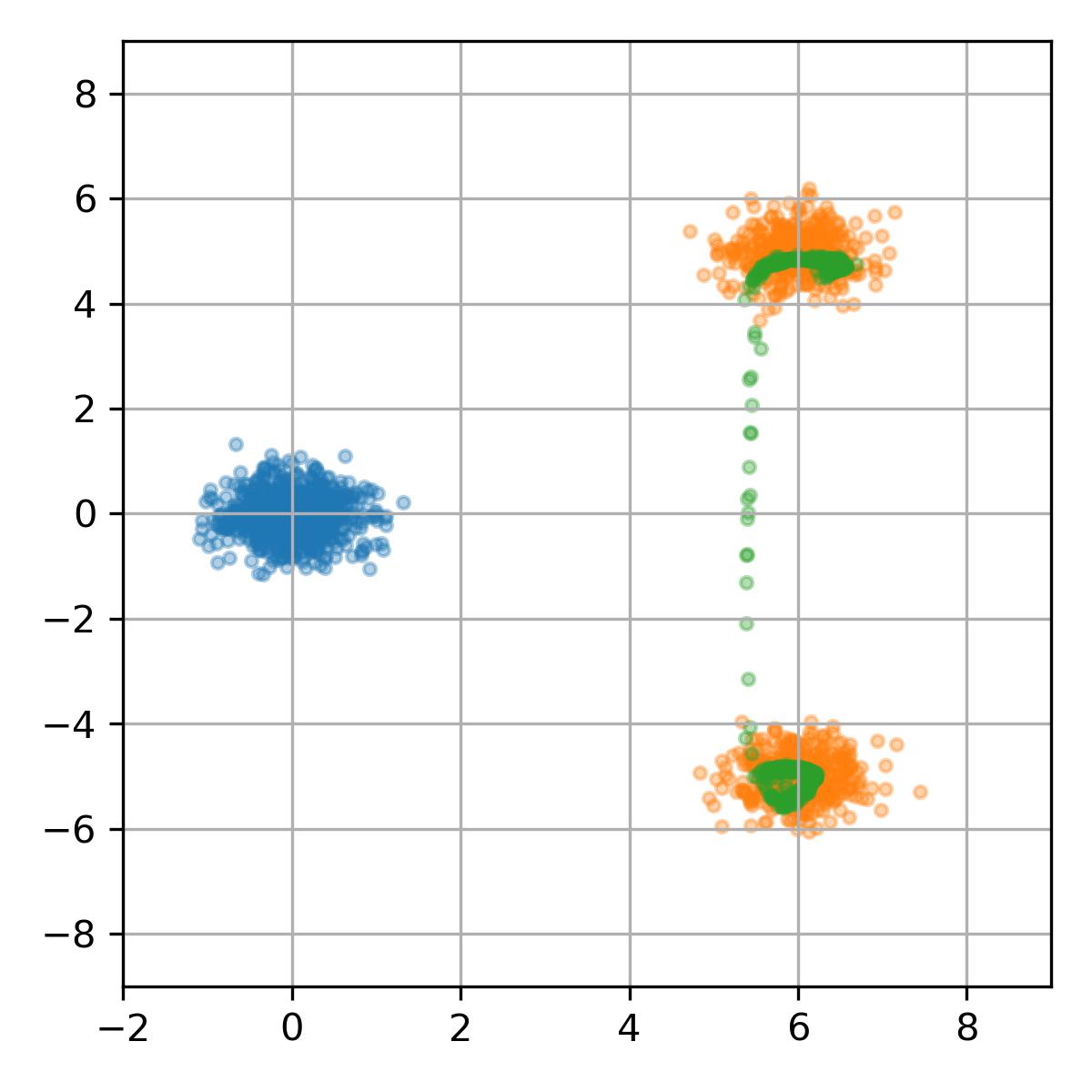}
        \caption*{(a) Vertical Gaussian}
    \end{minipage}
    \hfill
    \begin{minipage}[t]{0.46\linewidth}
        \centering
        \includegraphics[width=\linewidth]{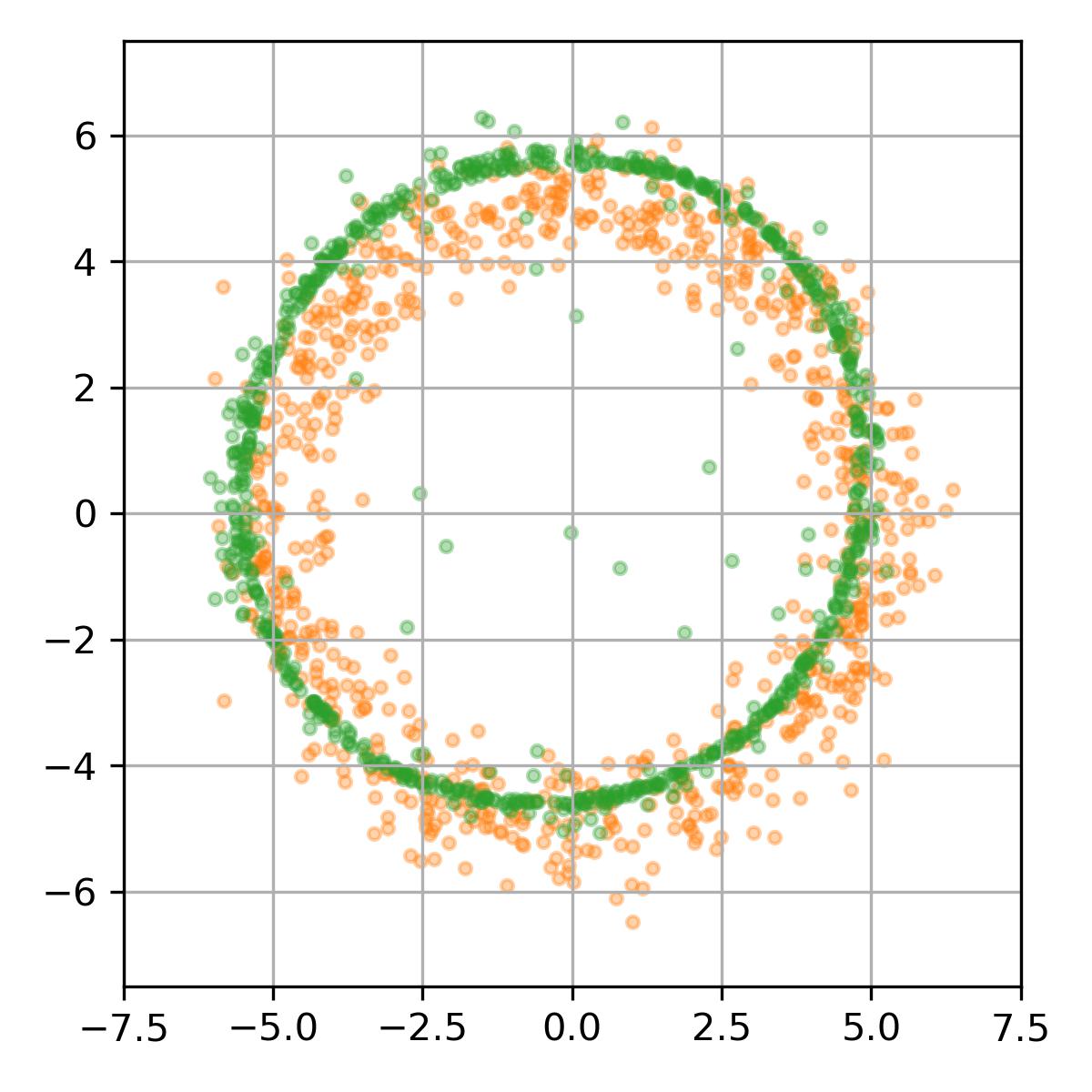}
        \caption*{(b) Ring distribution}
    \end{minipage}
    \caption{R3GAN on synthetic data examples. R3GAN still suffers from many-to-one collapse in both examples.}
    \label{fig:r3gan}
\end{figure}

\begin{figure*}[t]
    \centering
    \begin{minipage}[t]{0.46\linewidth}
        \centering
        \includegraphics[width=\linewidth]{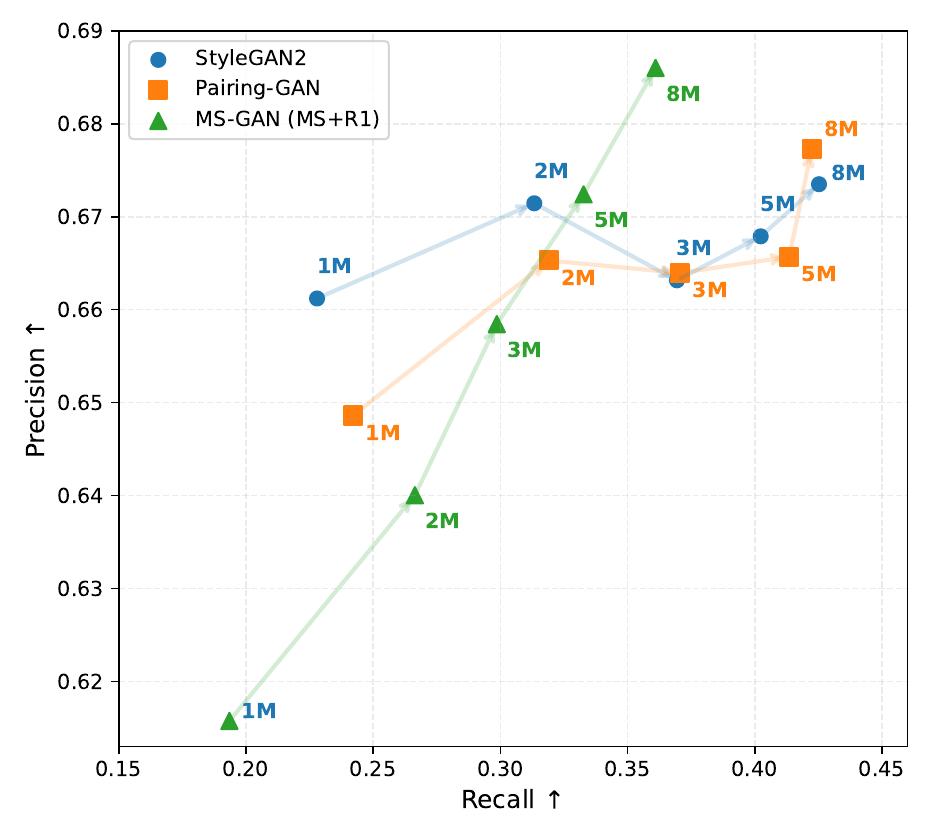}
        \caption*{(a) Precision--Recall trajectory}
    \end{minipage}
    \hfill
    \begin{minipage}[t]{0.46\linewidth}
        \centering
        \includegraphics[width=\linewidth]{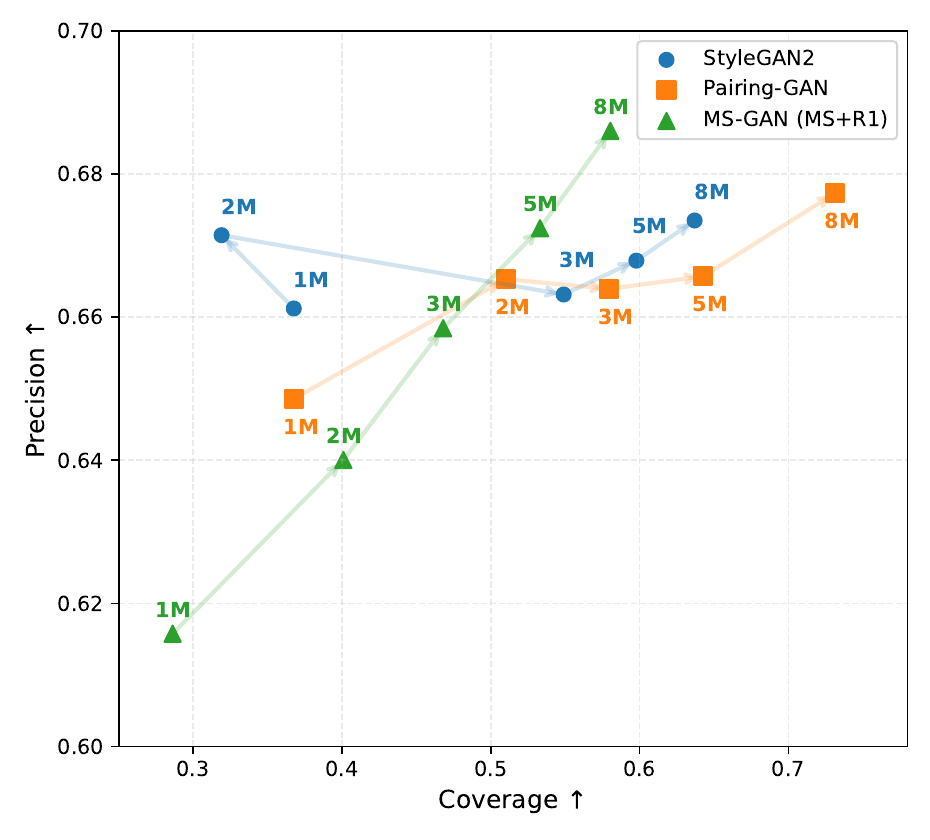}
        \caption*{(b) Precision--Coverage trajectory}
    \end{minipage}
    \caption{
        Training trajectories from 1M to 8M images without data augmentation. Results are averaged over three random seeds. Pairing-GAN yields competitive precision while maintaining similar recall (left), and achieves noticeably higher coverage (right), reducing many-to-one collapse effects in the no-aug, collapse-prone regime.
    }
    \label{fig:pr_pc_noaug_trajectory}
\end{figure*}
\begin{table*}[t]
\centering
\caption{
Precision, recall, and coverage on conditional CIFAR-10 without data
augmentation at 8M training images.
Results are averaged over three random seeds.
Coverage highlights differences in intra-class diversity that are not captured
by recall alone.
}
\label{tab:cifar_noaug_8m}
\small
\begin{tabular}{lcccc}
\toprule
Method & Precision $\uparrow$ & Recall $\uparrow$ & Coverage $\uparrow$ & FID $\downarrow$ \\
\midrule
StyleGAN2 & 0.673 & 0.425 & 0.637 & 9.219 \\
Pairing-GAN & 0.677 & 0.423 & \textbf{0.731} & \textbf{8.597}\\
MS-GAN (+R1) & 0.664 & 0.276 & 0.46 & 16.54\\
\bottomrule
\end{tabular}
\end{table*}
\subsection{Experimental Setup}

We evaluate the proposed pairing regularization on a diverse set of benchmarks,
ranging from low-dimensional toy distributions to high-dimensional image datasets.
Across all experiments, we mainly compare three methods: a Baseline GAN, which refers to a baseline GAN trained without additional regularization; MS-GAN, which incorporates mode-seeking regularization~\cite{mao2019mode}; and Pairing-GAN, which augments the baseline with the proposed pairing
regularization. Unless otherwise specified, all methods share the same generator and discriminator
architectures and are trained using identical optimization settings.
For image experiments, we adopt the StyleGAN2 training framework and apply regularizers such as $R_1$ penalty when stated explicitly.

We evaluate generative performance using multiple complementary metrics.
We report Fréchet Inception Distance (FID) to assess overall sample quality,
and precision and recall~\cite{sajjadi2018assessing,kynkaanniemi2019improved}
to characterize the trade-off between sample fidelity and distributional coverage.
In addition, we report the coverage metric proposed by
Sajjadi et al.~\cite{naeem2020prdc}, which measures the fraction of real
samples covered by the generated distribution and captures a notion of diversity that is distinct from recall. While recall quantifies whether generated samples can reach real data points, coverage emphasizes how uniformly the real data manifold is represented.
In particular, coverage is empirically sensitive to many-to-one collapse, where multiple latent codes map to overlapping regions of the data manifold, a failure mode that may not be reflected by recall alone. Moreover, we apply precision--recall (PR) and precision--coverage trajectories over training to visualize the learning dynamics and reveal differences in exploration and convergence behavior across methods. Unless otherwise specified, we report results averaged over three random seeds.

\subsection{Synthetic Data Experiments}

We first use controlled two-dimensional distributions to disentangle different failure modes of GAN training and to visually illustrate the effect of pairing regularization. All models in this section use simple multilayer perceptrons (MLPs) with two 128-unit hidden layers and LeakyReLU nonlinearities.

\paragraph{Vertical Gaussian Experiments}
First, we consider a two-dimensional Gaussian transport task to identify inter-mode and intra-mode collapse.
In this experiment, the prior distribution is a single Gaussian (shown in blue on the left), while the target distribution is a mixture of multiple Gaussians (shown in orange on the right).

In Figure~\ref{fig:2component_gaussian}(a), the standard GAN maps the prior to only a subset of the target mixture components, resulting in the bottom mode being poorly covered. This indicates clear inter-mode collapse, where parts of the target distribution are never reached. In Figure~\ref{fig:2component_gaussian}(b), adding a gradient penalty enables the generator to reach all target modes, indicating improved global coverage. Achieving a slight improvement on Recall (0.184 to 0.267). However, within each target mode, generated samples remain highly concentrated. Despite matching the target distribution at the marginal level, multiple latent inputs from the prior are mapped to nearly identical outputs, revealing persistent many-to-one (intra-mode) collapse.

In contrast, pairing regularization, as shown in Figure~\ref{fig:2component_gaussian}(c), promotes a more diverse mapping from the unimodal prior to the multimodal target.
Samples within each Gaussian component are distributed more uniformly, indicating improved intra-mode diversity and effective mitigation of many-to-one collapse.
However, the bottom target mode remains uncovered, suggesting that pairing regularization primarily influences local latent-to-data correspondence rather than global mode exploration. Finally, in Figure~\ref{fig:2component_gaussian}(d), combining pairing regularization with gradient penalty achieves both complete mode coverage and well-distributed samples within each mode. GAN+Pairing+R1 reaches 0.957 on Recall and 0.861 on Coverage. The corresponding quantitative results are shown in Table~\ref{tab:2component_pr}.

\begin{figure*}[t]
    \centering
    \begin{minipage}[t]{0.46\linewidth}
        \centering
        \includegraphics[width=\linewidth]{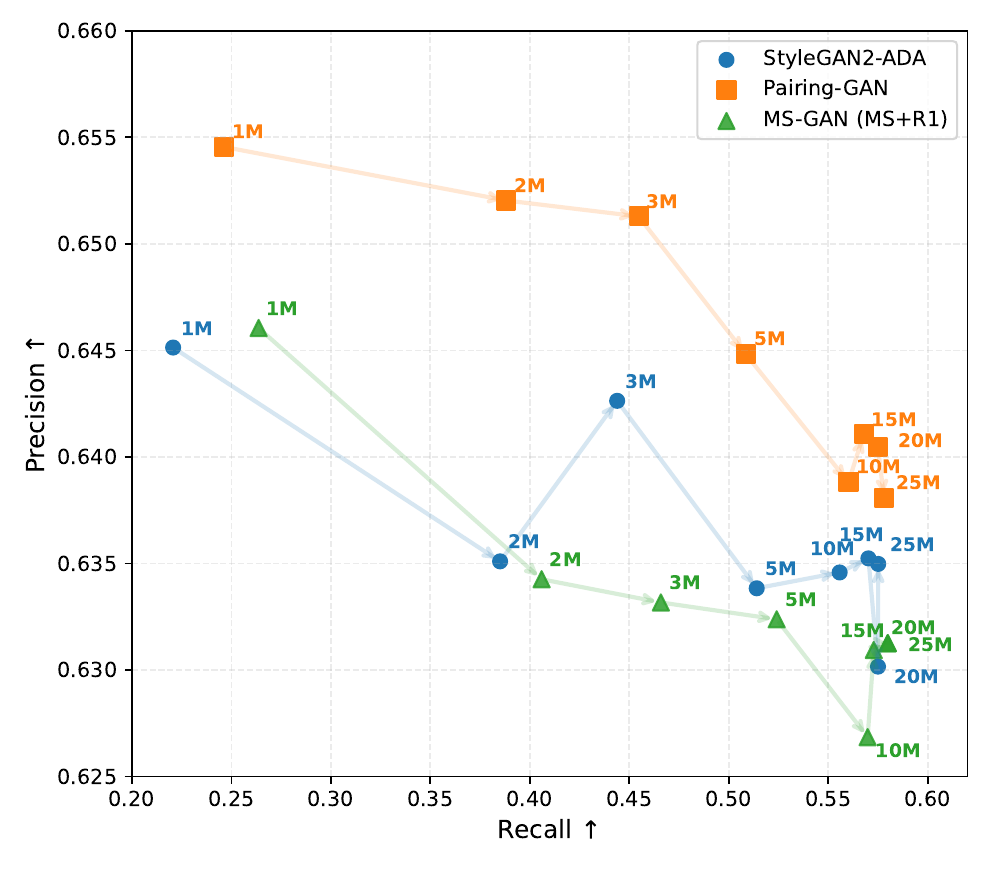}
        \caption*{(a) Precision--Recall trajectory}
    \end{minipage}
    \hfill
    \begin{minipage}[t]{0.46\linewidth}
        \centering
        \includegraphics[width=\linewidth]{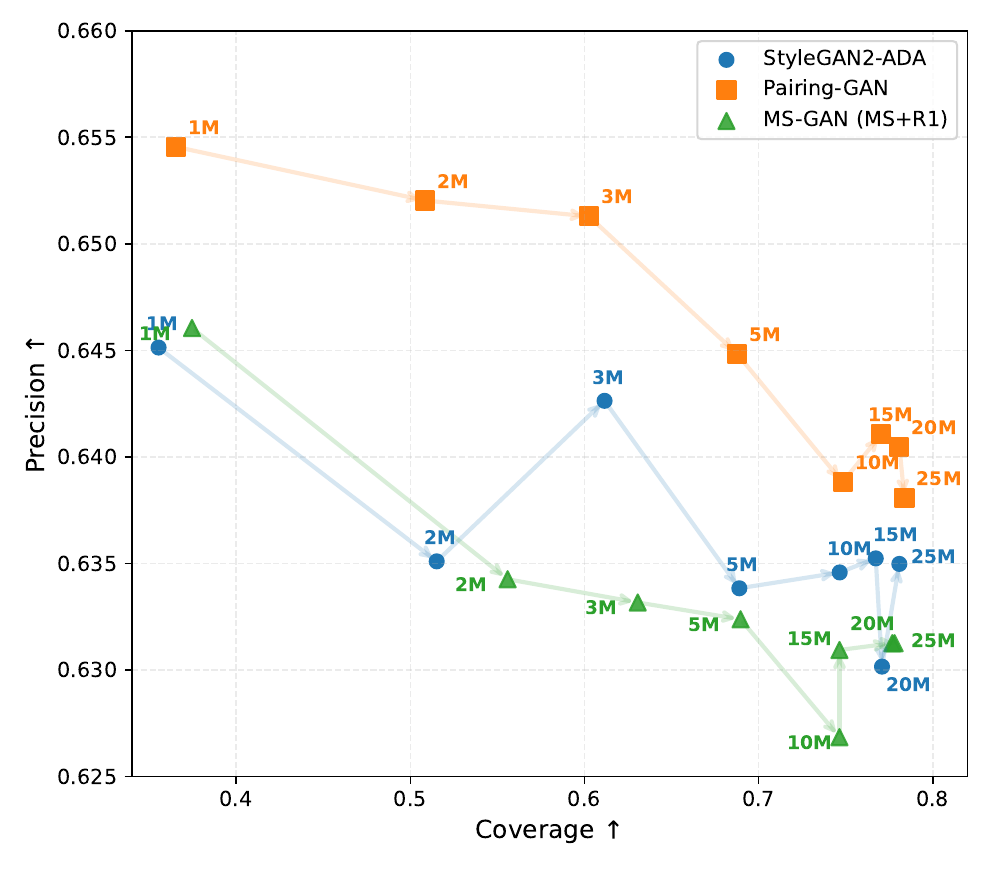}
        \caption*{(b) Precision--Coverage trajectory}
    \end{minipage}
    \caption{
        Training trajectories from 1M to 25M images with data argumentation. Results are averaged over three random seeds. Pairing-GAN and StyleGAN-ADA show highly similar trajectories in both Precision–Recall and Precision–Coverage spaces. Pairing-GAN achieves comparable recall and coverage, with marginally higher precision at each training stage.
    }
    \label{fig:pr_pc_aug_trajectory}
\end{figure*}
\begin{table*}[t]
\centering
\caption{Precision, recall, and coverage on conditional CIFAR-10 with adaptive
data augmentation at 25M training images.
Results are averaged over three random seeds.}
\small
\label{tab:appendix_cifar_aug_25m}
\begin{tabular}{lcccc}
\toprule
Method & Precision $\uparrow$ & Recall $\uparrow$ & Coverage $\uparrow$ & FID $\downarrow$ \\
\midrule
StyleGAN2 - ADA & 0.635 & 0.575 &  0.780 & 3.185\\
Pairing-GAN - ADA & 0.638 & 0.578 & \textbf{0.784} & \textbf{3.123}\\
MS-GAN(+R1) - ADA & 0.664 & 0.521 & 0.52 & 3.158\\
\bottomrule
\end{tabular}
\end{table*}

\paragraph{Ring Distribution}
We next consider a two-dimensional ring distribution, which represents a
continuous manifold with uniform density.
Compared to the Gaussian mixture, this setting removes explicit mode boundaries
and thus isolates many-to-one collapse from inter-mode dropping.

Based on the results in Figure~\ref{fig:ring} and Table~\ref{tab:ring_pr_coverage}, standard GAN and GAN with gradient penalty (R1) produce samples that concentrate on narrow arcs of the ring, indicating that multiple latent regions collapse
onto the same data neighborhoods.
Although R1 significantly improves recall, it fails to recover uniform coverage,
demonstrating that stabilization-based regularization alone does not prevent
many-to-one collapse and only achieve 0.372 on Coverage under high-recall regimes.

In contrast, pairing regularization explicitly encourages consistent
latent-to-data correspondence, resulting in a substantially more uniform distribution of samples along the ring. It achieves significant improvement in coverage (0.851) while maintaining high precision (0.970) and recall (0.962). In the high-recall setting, combining pairing with R1 yields results similar to GAN+Pairing.

\paragraph{From vanilla GAN to RpGAN}
We replace the vanilla GAN loss with the state-of-the-art relativistic loss~\cite{huang2024r3gan}. As shown in Figure~\ref{fig:r3gan}, the issue of many-to-one collapse can not be addressed by relativistic loss.

\subsection{Real Data Experiments}
We now turn to real-data experiments to examine whether many-to-one collapse and its associated dynamics persist in high-dimensional image generation. To isolate and analyze the many-to-one mode collapse, we evaluate our methods on conditional CIFAR-10 with and without data argumentation. 

\paragraph{Conditional CIFAR10 without ADA}
We first study conditional CIFAR-10 under the no-augmentation setting with a relatively strong gradient penalty, which constitutes a collapse-prone regime and allows training dynamics to be clearly observed by the precision--recall (PR) and precision--coverage training trajectories up to 8M training images in Figure~\ref{fig:pr_pc_noaug_trajectory}.

Pairing-GAN achieves the highest coverage 0.731 (compared with StyleGAN2 0.637) while maintaining comparable precision and recall to StyleGAN2. In contrast, MS-GAN (+R1) exhibits substantially reduced recall and coverage despite competitive precision, indicating severe collapse. These results highlight that coverage captures intra-class diversity that is not reflected in recall alone and confirm that pairing regularization effectively mitigates many-to-one collapse in the absence of data augmentation.

\paragraph{Conditional CIFAR-10 with ADA}
We next evaluate conditional CIFAR-10 under stabilized training with adaptive data augmentation (ADA) and a weaker gradient penalty. This setting reflects common modern GAN training practices and serves as a robustness check for pairing regularization in a data-augmentation-stabilized regime.

In Figure~\ref{fig:pr_pc_aug_trajectory}, we show the precision–recall and precision–coverage training trajectories from 1M to 25M images under data augmentation, averaged over three random seeds.
All three methods follow highly similar trajectories in both spaces, indicating that data augmentation largely stabilizes training and enforces comparable global coverage and local coverage on the original dataset. Pairing-GAN consistently attains marginally higher precision at corresponding training stages, while achieving nearly identical recall and coverage.

\section{Discussions}
Our work revisits mode collapse from a structural perspective and highlights many-to-one collapse as a distinct and practically relevant failure mode that is not well captured by standard marginal metrics. Unlike mode dropping, many-to-one collapse preserves global coverage while severely reducing intra-mode diversity, leading to generators that appear stable yet learn degenerate latent–data mappings.

Our experimental results suggest that commonly used stabilization and
regularization techniques primarily target inter-mode failures.
For example, gradient penalties and data augmentation are effective at
preventing severe mode dropping and improving global coverage, but do not
explicitly constrain how latent variables are mapped within each mode. Consequently, many-to-one collapse can persist even in settings where training appears stable and marginal metrics indicate good performance.

\paragraph{Combining gradient penalty and pairing}
Pairing regularization complements stabilization-based methods by explicitly encouraging a more faithful correspondence between latent variables and generated
samples. In collapse-prone regimes without data augmentation, where gradient penalties such as R1 primarily stabilizes discriminator training and prevents severe mode dropping, this additional structural constraint plays a critical role.
Our results show that pairing substantially improves intra-mode diversity, as reflected by higher coverage, while maintaining comparable precision and recall.
This suggests that while gradient penalties are effective at controlling inter-mode collapse, they do not explicitly prevent many-to-one mappings within individual modes.

\paragraph{ADA and pairing regularization}
For image datasets, where semantically meaningful data augmentations can be readily applied, ADA is an effective and practical tool for stabilizing GAN training. By exposing the discriminator to diverse local transformations of real samples,
ADA improves robustness, prevents severe mode dropping, and encourages both global coverage and local diversity in the generated samples.
In such settings, the overall training dynamics are already strongly regularized, leading to highly similar precision--recall and
precision--coverage trajectories across different methods.

Consequently, the improvement from pairing regularization with data augmentation is typically modest.
This behavior can be attributed to two factors. First, ADA-based training often employs relatively small gradient penalties (e.g.,
R1), which restricts the effective gradients propagated to the generator and limits the impact of additional regularizers. Second, data augmentation reshapes the effective data manifold observed during training and modifies the gradients the discriminator provides to the generator. Although evaluation is performed on the original data distribution, this implicit regularization already promotes both global coverage and intra-mode diversity, leaving limited room for further gains.

Importantly, this does not indicate that the fundamental many-to-one collapse problem is resolved.
Rather, the architectural tendency of GANs to learn degenerate many-to-one latent--data mappings remains, but is largely masked by augmentation-stabilized
training. This observation suggests that while ADA is highly effective for image datasets, diagnosing and mitigating many-to-one collapse requires complementary structural
regularization or evaluation protocols beyond augmentation alone.

\section{Conclusion}
This work revisits mode collapse in generative adversarial networks from a
structural perspective and identifies many-to-one collapse as a distinct and
practically relevant failure mode.
Unlike mode dropping, many-to-one collapse preserves global coverage while severely reducing intra-mode diversity, leading to generators that appear stable under standard metrics yet learn degenerate latent--data mappings. Through controlled toy experiments and image-generation benchmarks, we show that the many-to-one collapse can persist even under stabilized training regimes that employ commonly used techniques such as gradient penalties.

We propose pairing regularization as a simple and general mechanism to encourage more faithful latent--data correspondence. Our results demonstrate that pairing effectively mitigates many-to-one collapse in collapse-prone regimes, improving intra-mode diversity without sacrificing precision or recall.
In contrast, stabilization techniques such as gradient penalties and adaptive data augmentation primarily address inter-mode failures and training instability, and may mask underlying structural collapse rather than resolving it.

These findings highlight the importance of structural regularization and diagnostic evaluation beyond marginal distributions. We believe that explicitly addressing many-to-one collapse is essential for
developing generative models that not only match data distributions but also
learn meaningful and diverse latent representations.

% \begin{table*}[h]
% \centering
% \caption{Ablation study on \textbf{Stacked-MNIST} and \textbf{CIFAR-10}.
% For Stacked-MNIST, we report the number of covered modes, KL divergence to the uniform distribution,
% and precision. For CIFAR-10, we report FID for sample quality and Precision--Recall for diversity.
% Results on CIFAR-10 are averaged over 3 random seeds.}
% \label{tab:ablation_combined}
% \begin{tabular}{lcc|ccc}
% \toprule
% & \multicolumn{2}{c|}{\textbf{3-Stacked MNIST}} 
% & \multicolumn{3}{c}{\textbf{CIFAR-10}} \\
% \cmidrule(lr){2-3} \cmidrule(lr){4-6}
% \textbf{Method}
% & \textbf{Covered Modes} $\uparrow$
% & \textbf{KL} $\downarrow$
% & \textbf{FID} $\downarrow$
% & \textbf{Precision} $\uparrow$
% & \textbf{Recall} $\uparrow$ \\
% \midrule
% GAN + R1/R2
% & 984 / 1000 
% & 0.141
% & 3.99 %fid
% & 0.640
% & 0.547 \\

% + Pairing (Random)
% & xx / 1000
% & xx
% & xx
% & xx
% & xx \\

% + Pairing (Ours)
% & xx / 1000
% & xx
% & xx
% & xx
% & xx \\

% \textbf{Full Model}
% & \textbf{xx / 1000}
% & \textbf{xx}
% & \textbf{xx}
% & xx
% & \textbf{xx} \\
% \bottomrule
% \end{tabular}
% \end{table*}


% \bibliography{reference}

\bibliographystyle{icml2026}

%%%%%%%%%%%%%%%%%%%%%%%%%%%%%%%%%%%%%%%%%%%%%%%%%%%%%%%%%%%%%%%%%%%%%%%%%%%%%%%
%%%%%%%%%%%%%%%%%%%%%%%%%%%%%%%%%%%%%%%%%%%%%%%%%%%%%%%%%%%%%%%%%%%%%%%%%%%%%%%
% APPENDIX
%%%%%%%%%%%%%%%%%%%%%%%%%%%%%%%%%%%%%%%%%%%%%%%%%%%%%%%%%%%%%%%%%%%%%%%%%%%%%%%
%%%%%%%%%%%%%%%%%%%%%%%%%%%%%%%%%%%%%%%%%%%%%%%%%%%%%%%%%%%%%%%%%%%%%%%%%%%%%%%
\newpage
\appendix
\onecolumn

\section*{Appendix A. Additional Toy Experiments: 25-Gaussian Grid}
\begin{figure}[h]
    \centering

    \begin{minipage}[h]{0.23\textwidth}
        \centering
        \includegraphics[width=0.9\linewidth]{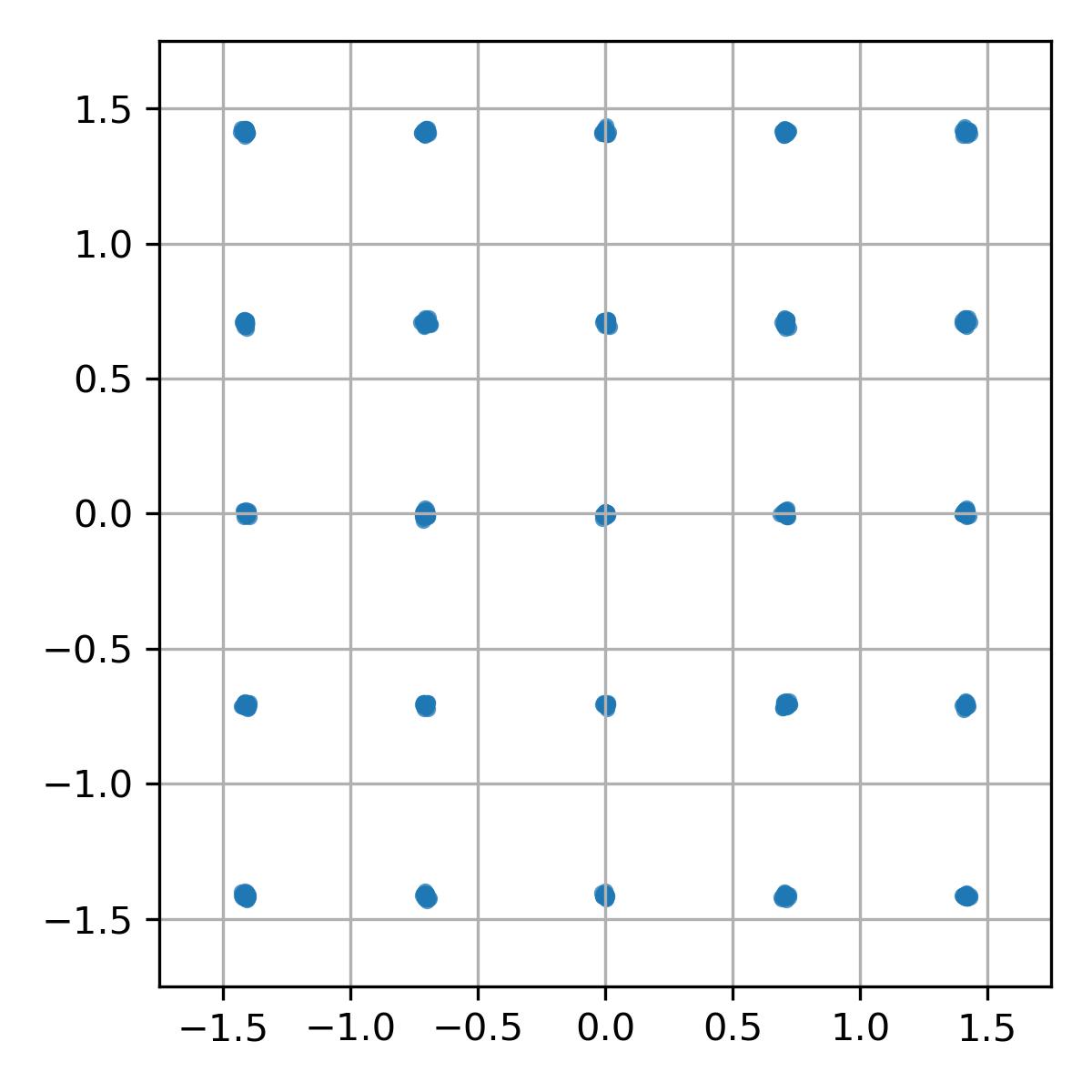}
        \\ (a) Target
    \end{minipage}
    \hspace{0.01\textwidth}
    \begin{minipage}[h]{0.23\textwidth}
        \centering
        \includegraphics[width=0.9\linewidth]{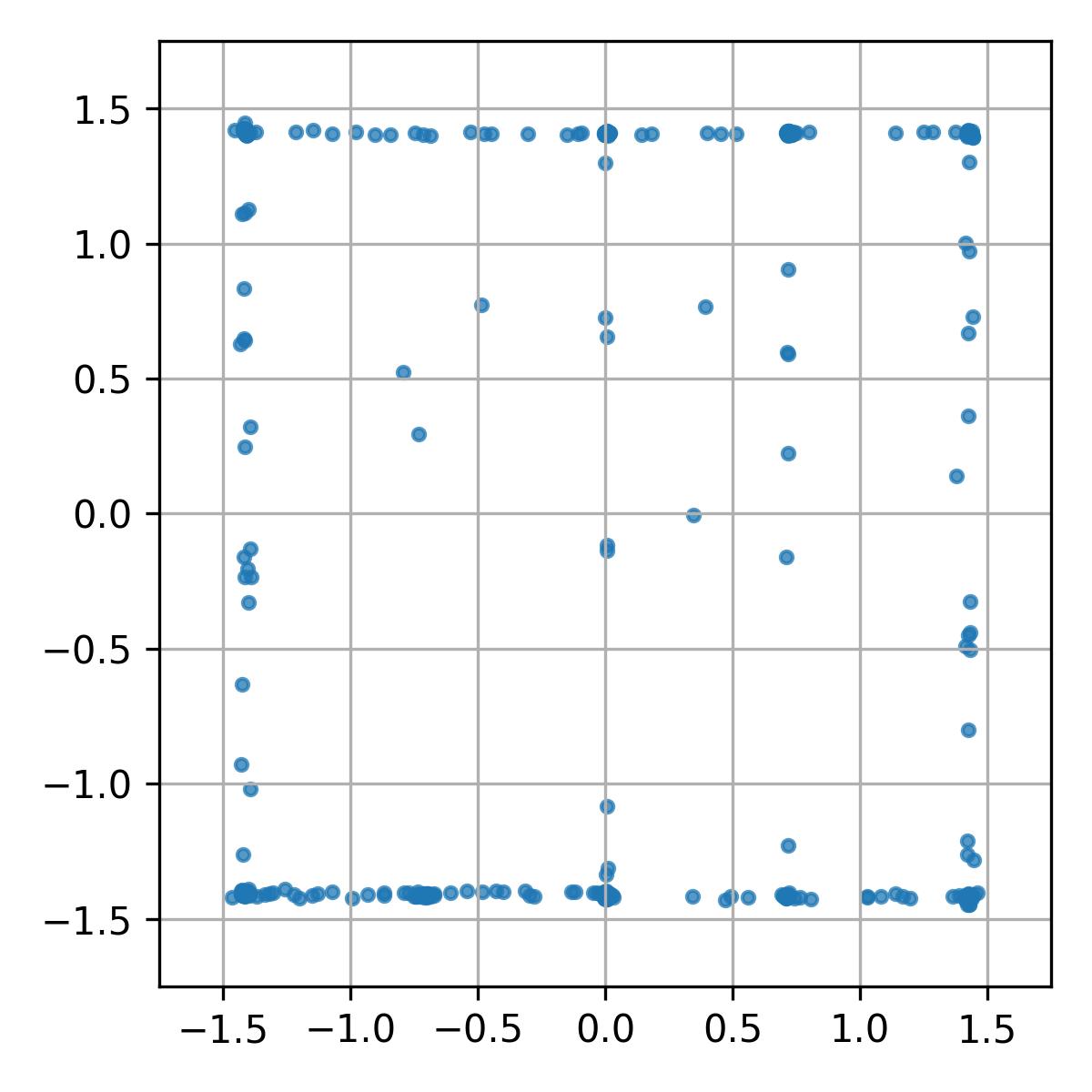}
        \\ (b) GAN
    \end{minipage}
    \hspace{0.01\textwidth}
    \begin{minipage}[h]{0.23\textwidth}
        \centering
        \includegraphics[width=0.9\linewidth]{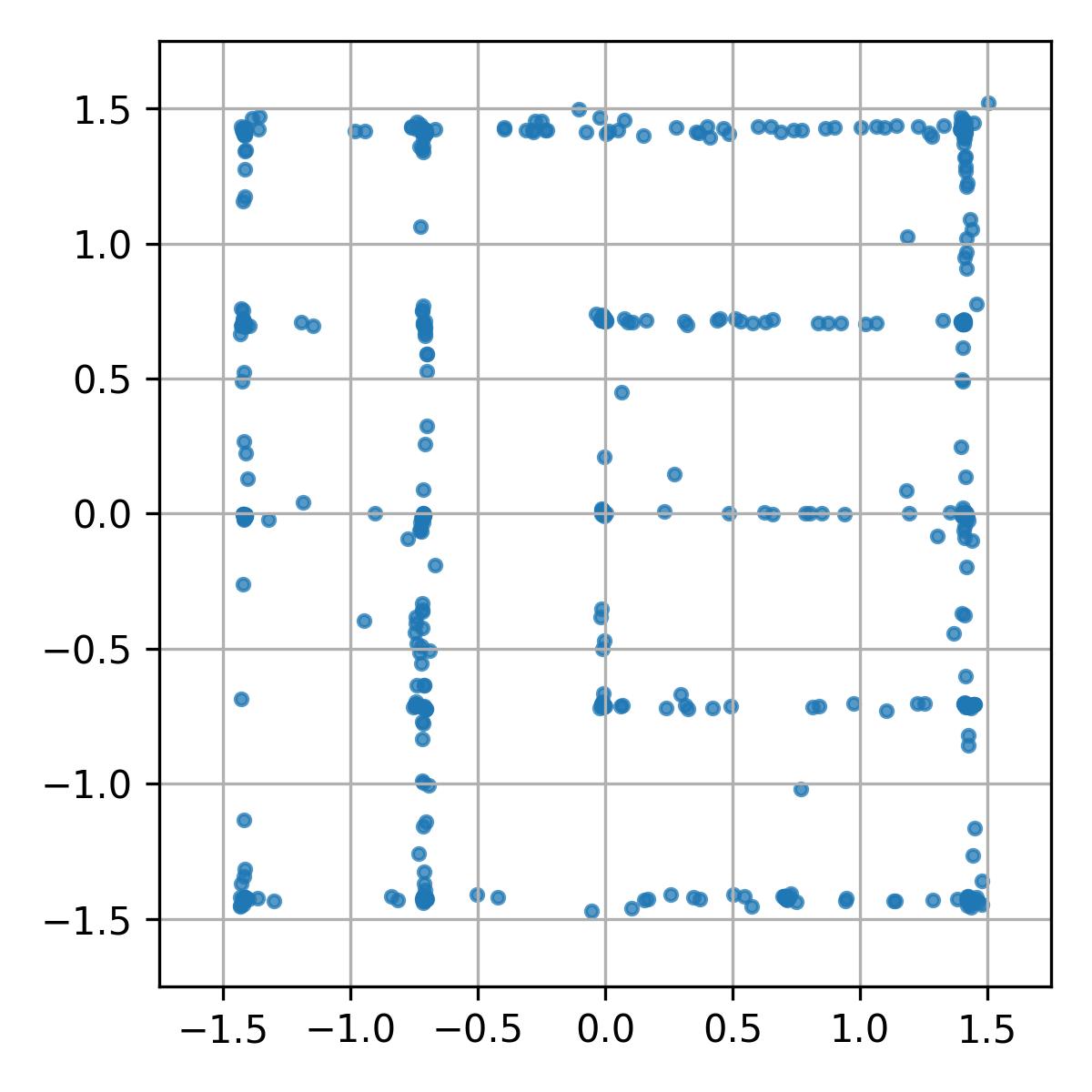}
        \\ (c) GAN + R1
    \end{minipage}
    \hspace{0.01\textwidth}
    \begin{minipage}[h]{0.23\textwidth}
        \centering
        \includegraphics[width=0.9\linewidth]{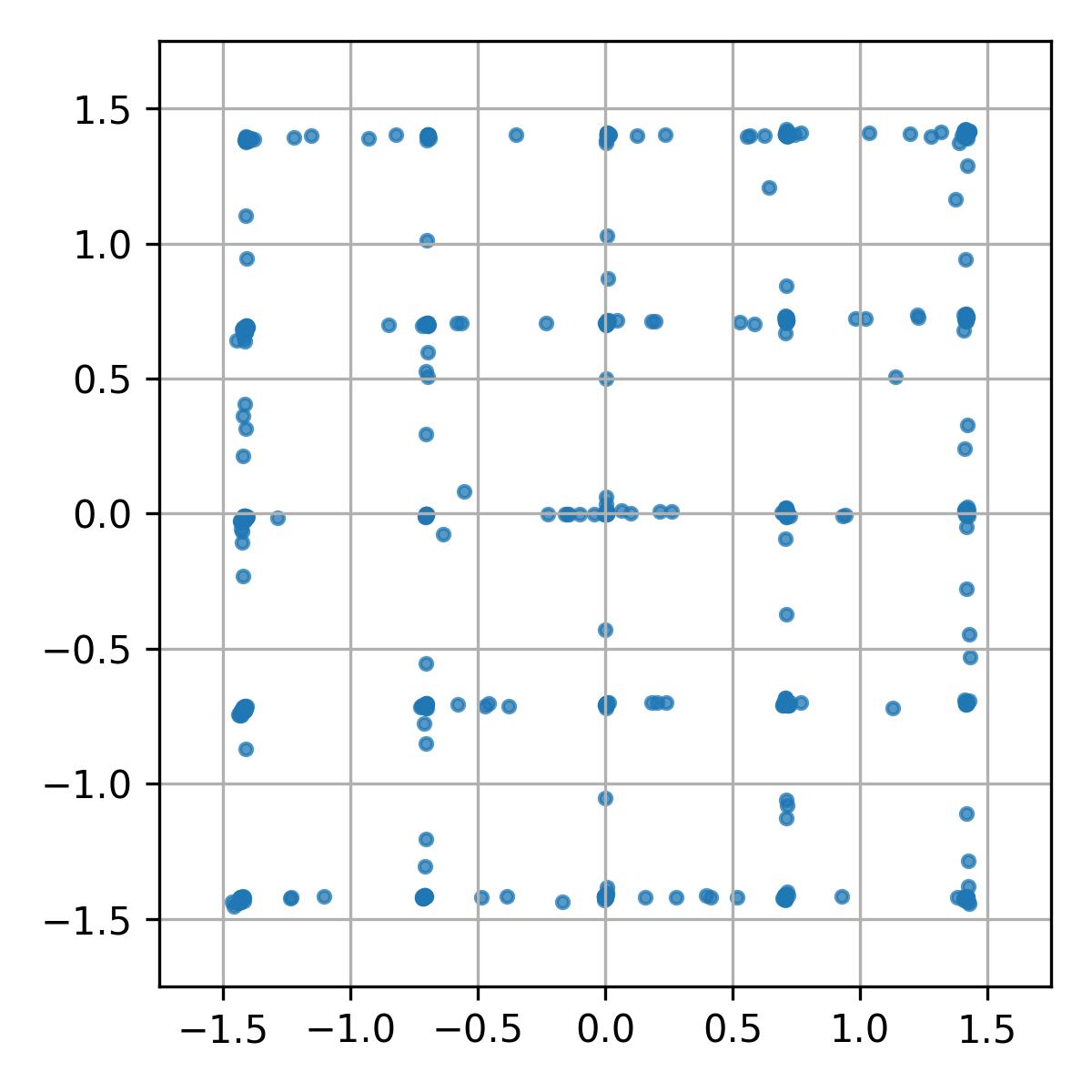}
        \\ (d) GAN + Pairing + R1
    \end{minipage}

    \caption{
        Toy 2D Gaussian grid task with 25 modes.
        Standard GAN and R1/R2 regularization suffer from many-to-one collapse, resulting in
        severe mode dropping and poor intra-mode coverage.
        Pairing regularization encourages latent--sample correspondence, leading to improved
        mode coverage and more uniform support across modes.
        }
    \label{fig:gaussian_grid}
\end{figure}
\begin{figure}[h]
    \centering

    \begin{minipage}{0.23\textwidth}
        \centering
        \includegraphics[width=0.9\linewidth]{25gauss_target.jpg}
        \\ (a) Target
    \end{minipage}
    \hspace{0.01\textwidth}
    \begin{minipage}{0.23\textwidth}
        \centering
        \includegraphics[width=0.9\linewidth]{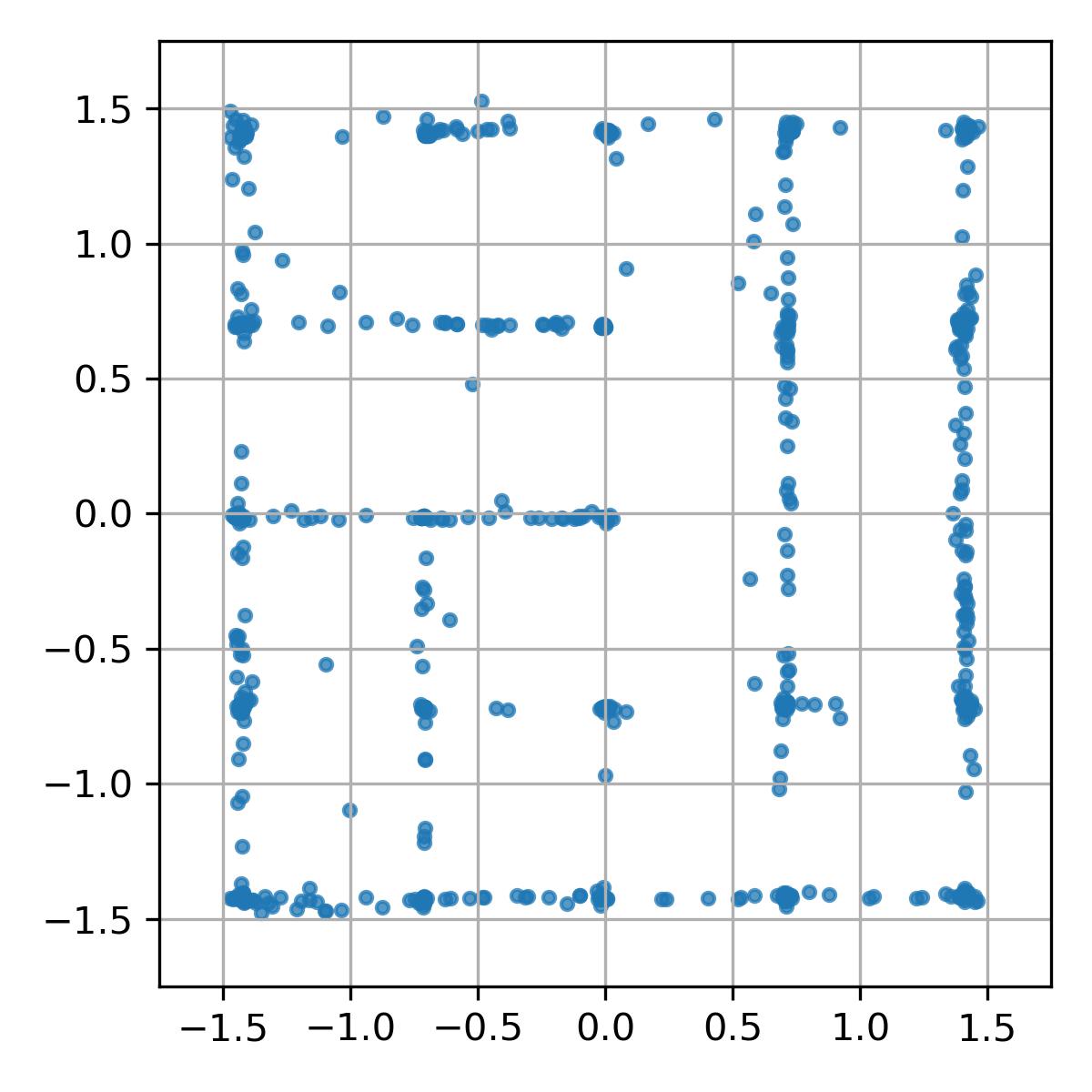}
        \\ (b) R3GAN
    \end{minipage}
    \hspace{0.01\textwidth}
    \begin{minipage}{0.23\textwidth}
        \centering
        \includegraphics[width=0.9\linewidth]{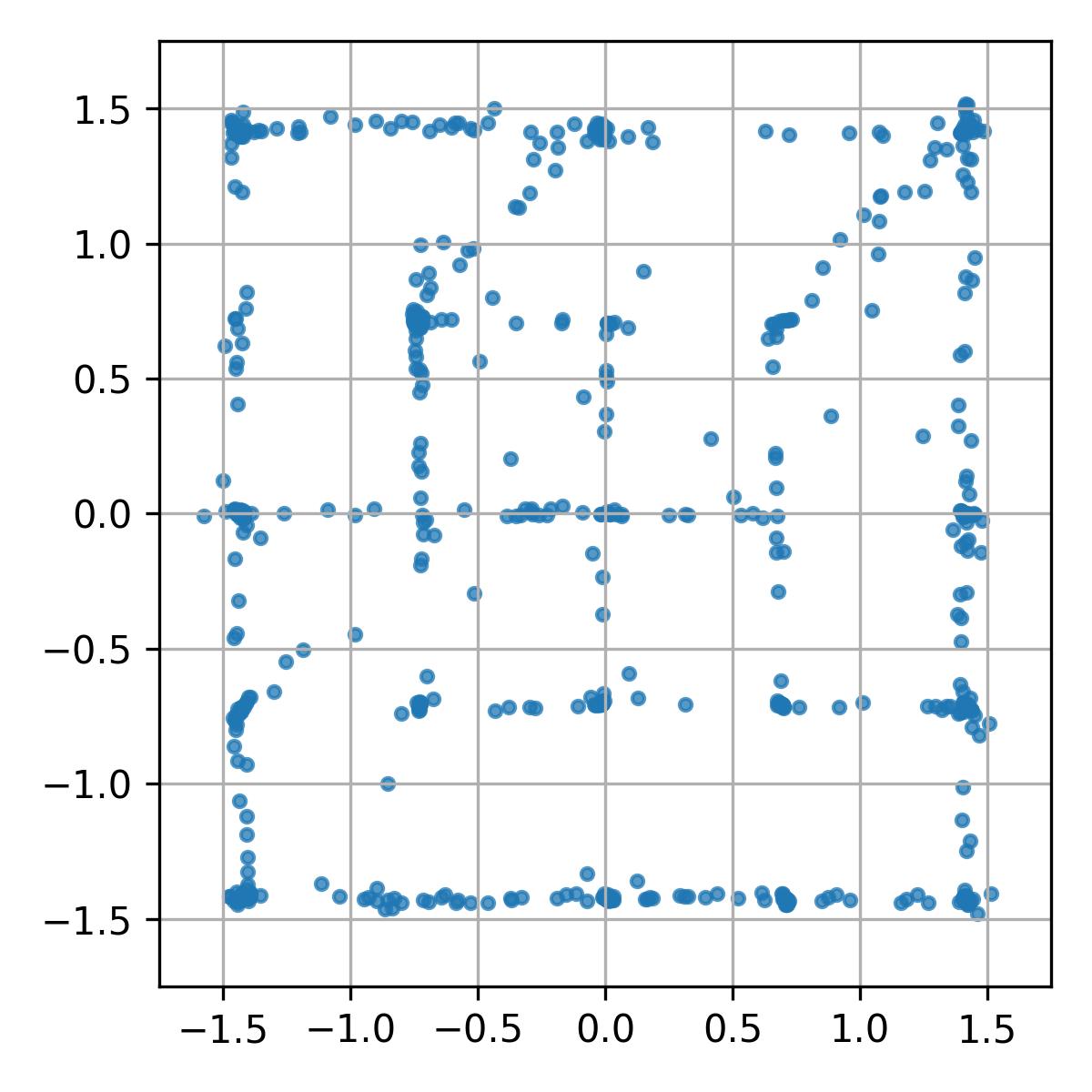}
        \\ (c) MS-GAN
    \end{minipage}
    \hspace{0.01\textwidth}
    \begin{minipage}{0.23\textwidth}
        \centering
        \includegraphics[width=0.9\linewidth]{25gauss_gan_pair_r1.jpg}
        \\ (d) Pairing-GAN
    \end{minipage}

    \caption{
        Toy 2D Gaussian grid task with 25 modes arranged in a 5x5 layout. (a) Target distribution. (b) R3GAN~\cite{huang2024r3gan}. (c) MS-GAN~\cite{mao2019mode}. (d) Pairing-GAN (ours).
        }
    \label{fig:gaussian_grid_different_methods}
\end{figure}
\begin{table}[h]
\caption{Quantitative results on the two-dimensional 25-component Gaussian grid.
Precision and recall measure sample fidelity and mode coverage, while
coverage captures the uniformity of the generator mapping across modes and
reveals many-to-one collapse.}
\label{tab:gaussian_grid_25}
\small
\centering
\begin{tabular}{lccc}
\toprule
Method & Precision $\uparrow$ & Recall $\uparrow$ & Coverage $\uparrow$ \\
\midrule
GAN                & 0.728 & \textbf{0.947} & 0.218 \\
GAN + R1           & 0.634 & 0.911 & 0.314 \\
GAN + Pairing + R1     & 0.759 & 0.940 & \textbf{0.432} \\
MS-GAN (+R1)     & 0.354 & 0.94 & 0.255 \\
\bottomrule
\end{tabular}
\end{table}
To examine whether many-to-one collapse persists in more complex multimodal settings, we consider the classical two-dimensional Gaussian grid with 25 modes. Compared to simple mixture or ring distributions, this setting introduces a denser and more structured mode configuration, enabling a more detailed inspection of latent-to-mode correspondences.

In Figure~\ref{fig:gaussian_grid}, all models achieve high recall, indicating that generated samples
lie close to the target manifold.
However, qualitative inspection reveals substantial differences in how latent variables are mapped to individual modes. Standard GANs and gradient-penalty-based variants tend to map large regions of the latent space to a subset of nearby modes, resulting in uneven coverage and many-to-one mappings within the grid. In contrast, pairing regularization encourages a more uniform allocation of latent variables across modes, leading to improved intra-mode diversity and more consistent coverage over the grid. These results confirm that many-to-one collapse is not limited to simple or
low-dimensional toy examples, but persists in structured multimodal
distributions, and that pairing remains effective in mitigating this failure
mode.

\section*{Appendix B. Additional CIFAR-10 Results}
We report additional quantitative results on conditional CIFAR-10 with and without adaptive data augmentation. Results are evaluated at 25M and 15M  training images to reflect fully converged
augmentation-stabilized training, and are averaged over three random seeds.
\begin{table}[h]
\centering
\caption{Precision, recall, and coverage on conditional CIFAR-10 without data
augmentation at 8M training images.
Results are averaged over three random seeds.}
\label{tab:appendix_cifar_noaug_8m}
\begin{tabular}{lcccc}
\toprule
Method & Precision & Recall & Coverage \\
\midrule
StyleGAN2 & 0.673 & 0.425 & 0.637  \\
Pairing-GAN ($\lambda{=}0.05$) & \textbf{0.677} & 0.422 & \textbf{0.731} \\
\bottomrule
\end{tabular}
\end{table}

\begin{figure}
    \centering
    \includegraphics[width=0.9\linewidth]{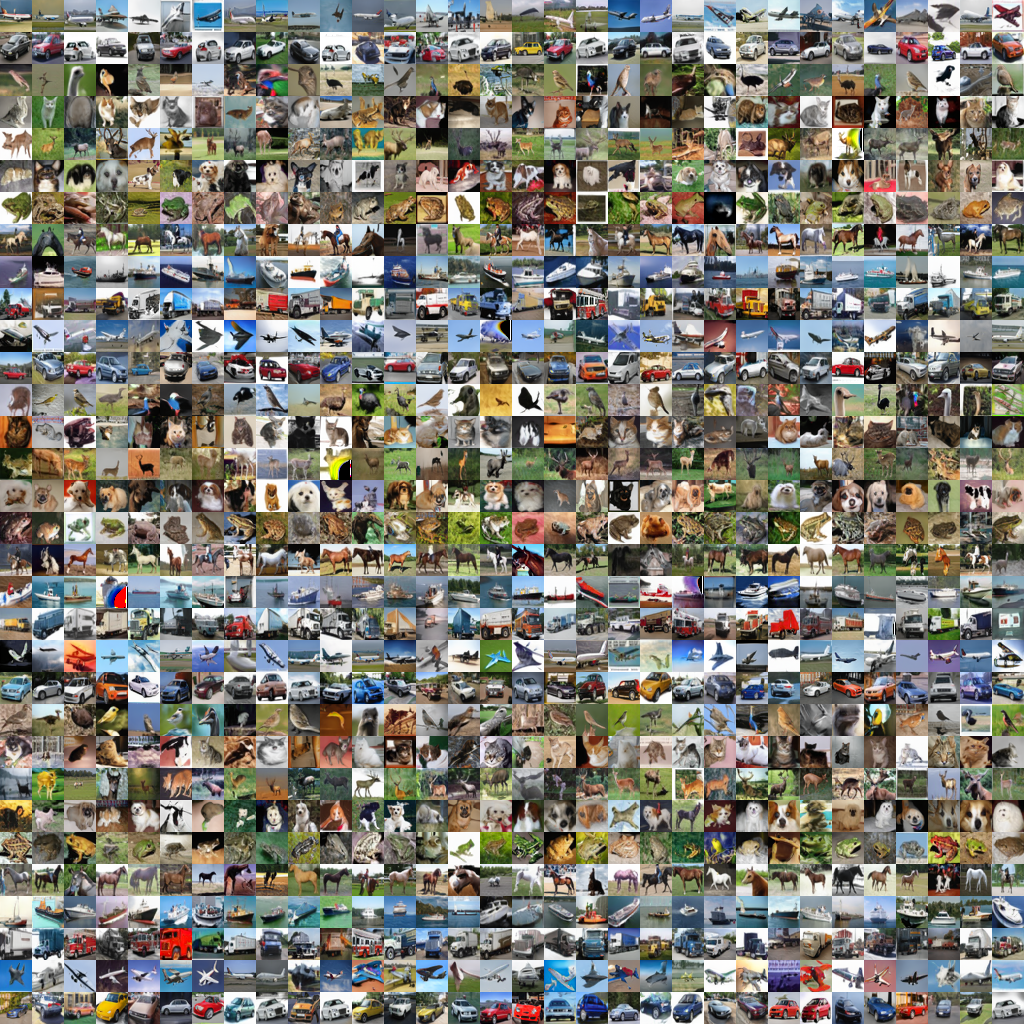}
    \caption{Pairing-GAN generated image on conditional CIFAR-10.}
    \label{fig:placeholder}
\end{figure}

\section*{Appendix C. Ablation Studies: Effect of Pairing Weight}

We examine the sensitivity of pairing regularization to the weighting parameter
$\lambda_{\text{pair}}$ under the no-augmentation setting.

\begin{table}[h]
\centering
\caption{Effect of pairing weight $\lambda_{\text{pair}}$ on coverage at 8M training images without data augmentation. $\lambda_{\text{pair}}=0.05$ provides a balanced trade-off.}
\label{tab:appendix_lambda}
\begin{tabular}{lcc}
\toprule
$\lambda_{\text{pair}}$ & Precision & Coverage \\
\midrule
0.01 & 0.6721 & 0.6812 \\
0.05 & \textbf{0.6773} & 0.7313 \\
0.10 & 0.6495 & \textbf{0.7364}  \\
\bottomrule
\end{tabular}
\end{table}
\section*{Appendix D. Implementation Details}

All experiments are conducted using the StyleGAN2 architecture with class
conditioning on CIFAR-10. Generators and discriminators follow the standard
configuration and are trained using Adam with default hyperparameters.
Training length is measured in the number of images seen by the discriminator. Pairing regularization is implemented using an auxiliary pairing network that maps latent variables and generated images into a shared embedding space, where
the pairing loss encourages matched latent--image pairs to have higher similarity than mismatched pairs.
The pairing weight is fixed to
$\lambda_{\text{pair}} = 0.05$ across all experiments. Regularization is applied using the R1 gradient penalty on real samples. In experiments without data augmentation, a larger R1 coefficient is used to stabilize training, whereas for experiments with data augmentation, adaptive data
augmentation (ADA) is employed following standard practice with a smaller R1 coefficient. Model performance is evaluated using precision, recall, and coverage metrics computed on generated and real samples, with results averaged over three random seeds. Unless otherwise specified, no-augmentation results are reported at 8M training
images, while augmentation-based results are reported at 25M training images to reflect converged training.

\end{document}